\documentclass[runningheads]{llncs}

\usepackage{eccv}

\usepackage{eccvabbrv}

\usepackage{graphicx}
\usepackage{booktabs}

\usepackage[accsupp]{axessibility}  

\usepackage{hyperref}

\usepackage{orcidlink}

\usepackage{microtype}
\usepackage{graphicx}
\usepackage{subcaption}
\usepackage{booktabs}
\usepackage{enumitem}
\usepackage{listings}
\usepackage{amsmath,amssymb}
\usepackage{tabularx}
\usepackage{xspace}
\usepackage{makecell}
\usepackage{array}
\usepackage{float}
\usepackage{stfloats}
\usepackage{xcolor}
\usepackage{wrapfig}
\usepackage{tcolorbox}
\usepackage{multirow}
\usepackage{pifont}
\tcbuselibrary{skins,breakable,listings}

\begin{document}

\title{BrainMem: Brain-Inspired Evolving Memory for Embodied Agent Task Planning} 

\titlerunning{BrainMem: Brain-Inspired Evolving Memory}

\author{
Xiaoyu Ma\inst{1} \and
Lianyu Hu\inst{2} \and
Wenbing Tang\inst{1} \and
Zixuan Hu\inst{1} \and
Zeqin Liao\inst{1} \and
Zhizhen Wu\inst{1} \and
Yang Liu\inst{1}
}

\authorrunning{X. Ma,  et al.}

\institute{
Nanyang Technological University, Singapore
\email{ma0009yu@e.ntu.edu.sg, wenbing.tang@ntu.edu.sg, ZIXUAN014@e.ntu.edu.sg, zeqin.liao@ntu.edu.sg, c250082@e.ntu.edu.sg, yangliu@ntu.edu.sg}
\and
Tianjin University, Tianjin, China
\email{hly2021@tju.edu.cn}
}

\maketitle

\begin{abstract}
  Embodied task planning requires agents to execute long-horizon, goal-directed actions in complex 3D environments, where success depends on both immediate perception and accumulated experience across tasks. However, most existing LLM-based planners are stateless and reactive, operating without persistent memory and therefore repeating errors and struggling with spatial or temporal dependencies. We propose \textbf{BrainMem} (Brain-Inspired Evolving Memory), a training-free hierarchical memory system that equips embodied agents with working, episodic, and semantic memory inspired by human cognition. BrainMem continuously transforms interaction histories into structured knowledge graphs and distilled symbolic guidelines, enabling planners to retrieve, reason over, and adapt behaviors from past experience without any model fine-tuning or additional training. This plug-and-play design integrates seamlessly with arbitrary multi-modal LLMs and greatly reduces reliance on task-specific prompt engineering. Extensive experiments on four representative benchmarks, including EB-ALFRED, EB-Navigation, EB-Manipulation, and EB-Habitat, demonstrate that BrainMem significantly enhances task success rates across diverse models and difficulty subsets, with the largest gains observed on long-horizon and spatially complex tasks. These results highlight evolving memory as a promising and scalable mechanism for generalizable embodied intelligence.
\end{abstract}

\section{Introduction}
\label{sec:intro}

Embodied AI seeks to develop intelligent autonomous agents capable of executing complex tasks through situated interaction with dynamic physical environments~\cite{liu2025aligning,liu2025embodied}. Given the dynamic nature of embodied environments and the long-horizon nature of practical tasks, agents require planning capabilities to decompose abstract instructions into executable action sequences~\cite{hu2025flare,fei2025unleashing,zhang2025lamma}. Embodied task planning provides this fundamental capability by generating feasible action plans from high-level instructions through reasoning about spatial relations, object affordances, action preconditions, and environmental dynamics~\cite{zou2025embodiedbrain}.

\begin{figure}[tb]
    \centering
    \scriptsize
    \textbf{Failure Case 1: \textit{``Place a cleaned sponge in a bathtub''}} \\
    \begin{tabular}{ccc}
        \includegraphics[width=0.30\linewidth]{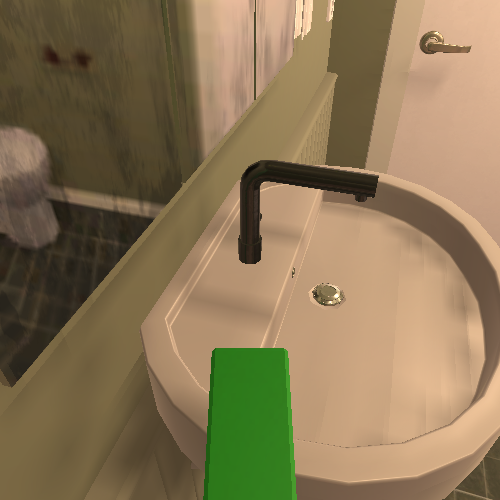} &
        \includegraphics[width=0.30\linewidth]{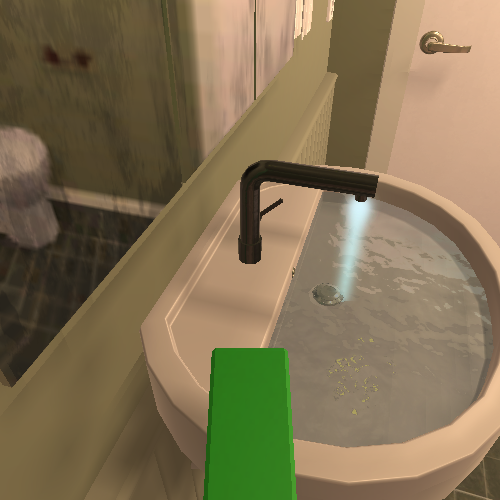} &
        \includegraphics[width=0.30\linewidth]{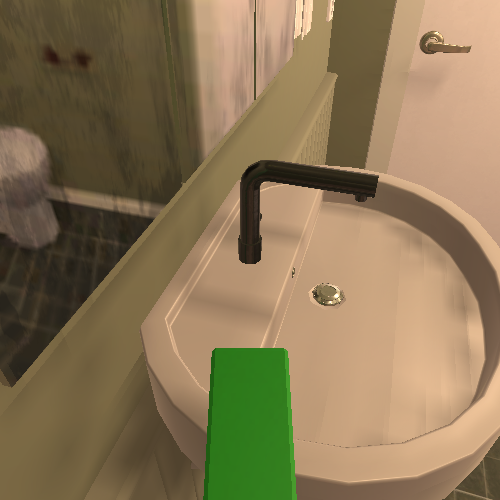} \\
        \small Find a faucet & \small Turn on faucet & \small Turn off faucet \\
    \end{tabular}

    \raggedright
    {\footnotesize \textit{Failure reason: The agent loses short-term context and keeps toggling the faucet, unaware the sponge is still held.}} \\[0.3em]

    \centering
    \textbf{Failure Case 2: \textit{``Place a glass cup with butter knife on the table''}} \\
    \begin{tabular}{ccc}
        \includegraphics[width=0.30\linewidth]{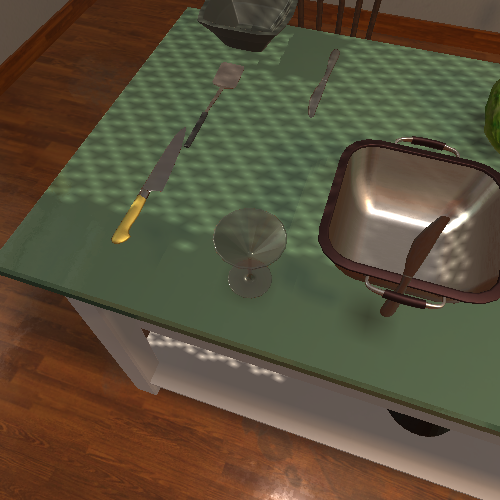} &
        \includegraphics[width=0.30\linewidth]{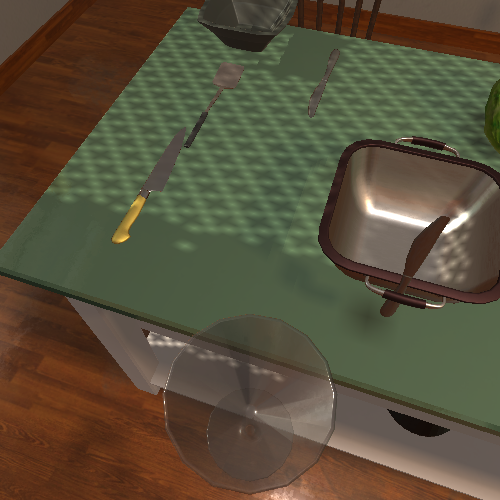} &
        \includegraphics[width=0.30\linewidth]{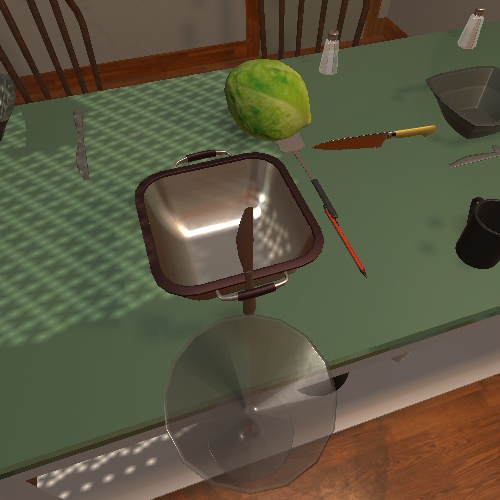} \\
        \small Find a cup & \small Pick up cup & \small Find a butter knife \\
    \end{tabular}

    \raggedright
    {\footnotesize \textit{Failure reason: The agent lacks experience combining objects, failing to place the knife into the cup and looping actions.}} \\
    \vspace{-0.25cm}
    \caption{\textbf{Typical failure modes of stateless planners.}
    (Top) The agent repeatedly toggles the faucet without completing the washing step.
    (Bottom) The agent fails to compose “pick up” and “place” actions due to missing object-combination experience.}
    \label{fig:intro_failures}
    \vspace{-3.3em}
\end{figure}

Existing embodied task planning approaches predominantly leverage Large Language Models (LLMs) as planning engines, exploiting their commonsense knowledge to generate action plans~\cite{song2023llm,glocker2025llm}. However, these LLM-based planners lack effective memory mechanisms to maintain task context and accumulate experiential knowledge. 
As illustrated in the first case of Fig.~\ref{fig:intro_failures}, when tasked with cleaning a sponge, an LLM-based agent opens the faucet but fails to place the sponge under water, instead repetitively toggling the faucet due to insufficient short-term memory to track subgoal completion. 
Beyond such short-term memory deficits, these planners also lack enough long-term experiences. As shown in the second case of Fig.~\ref{fig:intro_failures}, when the agent is asked to place a glass cup with a butter knife on the table, it doesn't realize the relationships between different objects and thus fail to correctly compose the order of various actions.

Despite the impressive reasoning ability of LLMs, their stateless nature prevents them from accumulating knowledge across tasks or maintaining coherent multi-step contexts within a single episode~\cite{zhao2024expel}. This results in agents that behave reactively rather than reflectively, which are unable to recall past outcomes or refine strategies over time. In contrast, human cognition relies on hierarchical memory systems that operate at different timescales: working memory for short-term situational awareness, episodic memory for structured recollection of experiences, and semantic memory for abstracted knowledge and generalization~\cite{zhang2025survey,gazzaniga2006cognitive,sumers2023cognitive}. Critically, these systems dynamically interact and evolve, enabling humans to learn from experience and progressively improve task performance~\cite{gazzaniga2006cognitive}.

To this end, we introduce \textbf{BrainMem} (\textbf{Brain-Inspired Evolving Memory}), a plug-and-play, training-free memory framework that augments diverse multi-modal LLM planners for human-like recall and adaptation.
BrainMem comprises three complementary modules operating at distinct temporal scales: a \textbf{working memory} that buffers recent actions, feedback, and agent states for stable short-horizon reasoning; a \textbf{episodic memory} that encodes task trajectories as action–state and spatial–object relations for long-horizon recall; and a \textbf{semantic memory} that abstracts successful patterns, failure causes, and distilled guidelines for future reuse.
These memory components evolve through continuous interaction. During execution, working memory updates with each action while episodic graphs accumulate trajectory structures. After task completion, episodic experiences consolidate into semantic guidelines through automated summarization, enabling progressive improvement without gradient updates or external supervision. 
During planning, BrainMem retrieves contextually relevant experiences to enhance decision-making quality.

We evaluate BrainMem on four diverse embodied planning benchmarks: EB-ALFRED, EB-Navigation, EB-Manipulation and EB-Habitat. Experimental results demonstrate that BrainMem significantly improves task success rates compared to baselines, with particularly substantial gains on long-horizon tasks requiring multi-step coordination and compositional reasoning. Ablation studies further validate the necessity of hierarchical memory organization and evolving mechanisms for effective embodied planning.

In summary, our key contributions are threefold:
\begin{itemize}[leftmargin=*, itemsep=0pt, topsep=1pt]
    \item A brain-inspired evolving memory architecture that unifies working, episodic, and semantic memory for dynamic reasoning and experience accumulation in embodied agents.
    \item Adaptive memory consolidation and retrieval mechanisms that enable plug-and-play integration with diverse  planners, requiring no task-specific fine-tuning.
    \item 
     Extensive experiments and analyses showing that BrainMem substantially improves task success rates across multiple embodied benchmarks with strong generalizability.
\end{itemize}

\section{Related Work}

\subsection{Embodied Task Planning}
Embodied task planning requires agents to decompose high-level instructions into executable action sequences through grounded reasoning in interactive environments. 
Early approaches employed symbolic planners (e.g., PDDL)~\cite{jiang2019task,zhang2025lamma} or reinforcement learning methods~\cite{hu2025flare,fei2025unleashing}, which either required extensive manual domain engineering or demanded large-scale training data, limiting their applicability to open-ended embodied scenarios.
The emergence of LLMs marked a paradigm shift, enabling zero-shot or few-shot task planning through natural language grounding. SayCan~\cite{ahn2022can} pioneered the integration of language priors with affordance functions, grounding abstract instructions in physically executable actions. Inner Monologue~\cite{huang2022inner} introduced closed-loop feedback mechanisms, allowing agents to revise plans based on success detection and environmental observations. Subsequent work enhanced reasoning through structured prompting strategies: ReAct~\cite{yao2022react} interleaved reasoning traces with action execution, while Reflexion~\cite{shinn2023reflexion} enabled self-reflection and error correction within episodes through verbal feedback. 
However, these methods operate without persistent memory mechanisms, treating each episode independently and failing to accumulate knowledge across tasks. Our work addresses this limitation by introducing a brain-inspired, hierarchical memory architecture that enables experience retention and progressive strategy refinement without gradient-based retraining.

\subsection{Memory for Embodied Agents.}
Memory mechanisms play a crucial role in enabling agents to maintain context and leverage past experiences in embodied task planning. Early methods such as model-free episodic control~\cite{blundell2016model} and neural episodic control~\cite{pritzel2017neural} stored high-value trajectories for rapid adaptation via reinforcement learning. Differentiable memory systems like Neural Turing Machines and Differentiable Neural Computers~\cite{graves2014ntm,graves2016dnc} provided learnable mechanisms to store and retrieve long-range temporal contexts for improved decision-making. For embodied navigation, spatially grounded memories such as Neural Map~\cite{parisotto2017neural} and graph-based semantic mapping~\cite{chaplot2020learning,zhang2021semanticnav} were introduced to encode egocentric structures for exploration. 

With the rise of LLM-based planning, recent frameworks store trajectory logs, object graphs, or textual summaries to provide planning context~\cite{lei2025stma,ginting2025enter,mao2025meta}. Recent efforts have further incorporated structured 3D scene memory into LLM-based embodied systems. 3D-Mem~\cite{yang2024threedmem} introduces a spatial memory module that maintains persistent 3D scene representations for embodied reasoning, enabling agents to accumulate geometric and semantic information. Similarly, 3DLLM-Mem~\cite{hu2025threedllmmem} integrates large language models with structured 3D memory to enhance long-horizon task execution and spatial grounding. RoboMemory~\cite{lei2025robomemory} advanced this direction by organizing episodic, semantic, and spatial memories for closed-loop planning. However, existing approaches typically lack task-specific working memory during execution or mechanisms for within-episode experience refinement. Our BrainMem addresses these limitations by maintaining explicit short-term context, representing episodic traces as dual knowledge graphs, and introducing a guided-retry mechanism that enriches experiences before consolidation.

\subsection{Brain-Related Methodologies.}
The human brain employs sophisticated hierarchical memory architectures that enable continuous learning and adaptive decision-making. Cognitive neuroscience has identified three distinct yet interconnected memory systems~\cite{baddeley2012working,tulving1983elements}: working memory maintains immediate task context and active goals during execution, episodic memory stores specific experiences with rich temporal and spatial details, and semantic memory contains abstracted, generalized knowledge applicable across contexts. 
These principles have inspired various AI architectures. HippoRAG~\cite{jimenez2024hipporag} implements hippocampal indexing via knowledge graphs, Generative Agents~\cite{park2023generative} consolidate experiences through reflection, and MemGPT~\cite{packer2023memgpt} separates working from long-term memory.
However, these approaches primarily target conversational or knowledge retrieval tasks and lack embodied-specific mechanisms such as spatial grounding, action-outcome dependency modeling, and real-time execution feedback integration.
Our BrainMem addresses this gap by adapting brain-inspired memory consolidation and retrieval mechanisms specifically for embodied planning, enabling agents to maintain context, learn from experiences, and progressively refine strategies.

\section{Method}
\label{sec:method}

\begin{figure*}[t]
    \centering
    \includegraphics[width=1\linewidth]{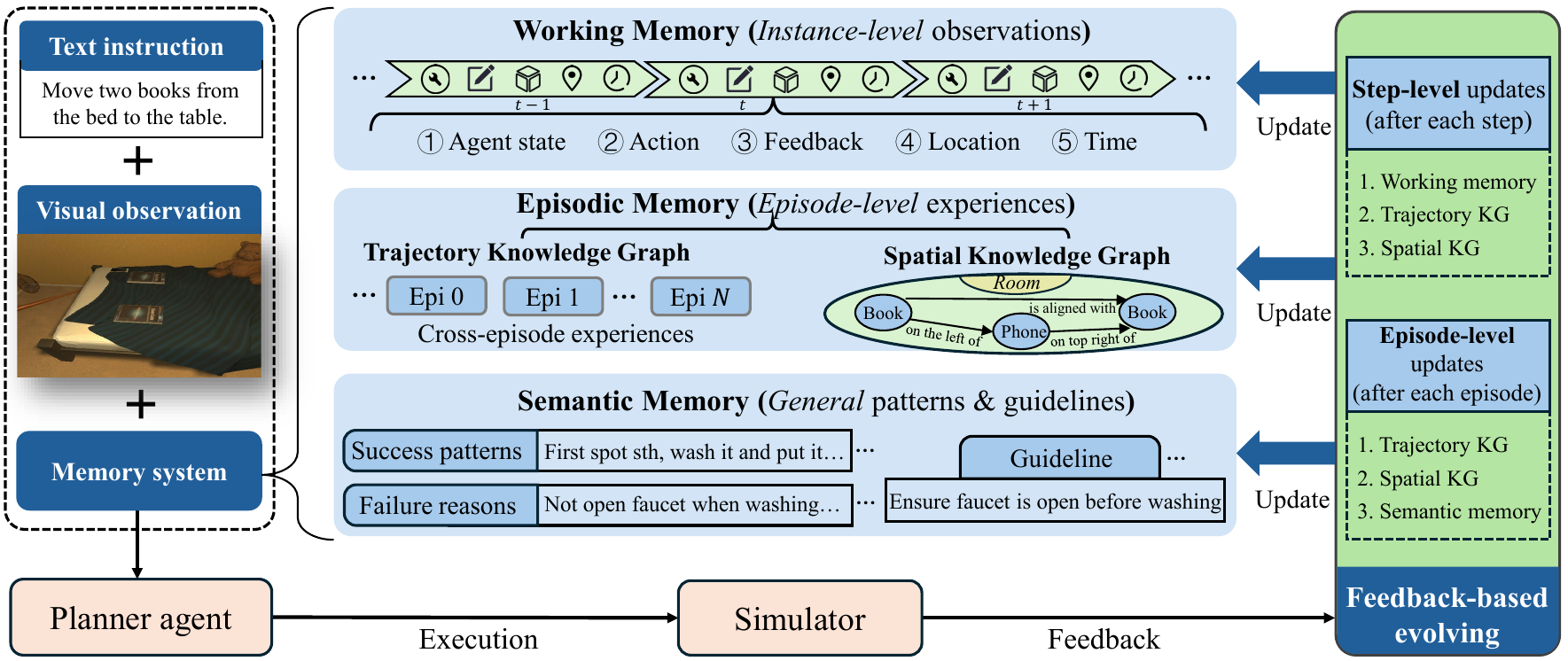}
    \caption{\textbf{Overall framework of the proposed BrainMem system.}
    At execution time (left), the \textit{Experience Agent} accumulates memory across three levels: Working Memory (short-term), Episodic Memory (task-specific), and Semantic Memory (task-agnostic, long-term insights). At inference (right), the \textit{Planner Agent} retrieves relevant memory to guide goal-directed action generation.}
    \label{fig:memory_framework}
\end{figure*}

\subsection{Overview}
Embodied task planning involves an agent situated in a dynamic 3D environment, which perceives the environment, reasons over robot states, and executes actions to complete multi-step goals described by natural language instructions such as “put the apple in the fridge”~\cite{alfred}. This complex task requires \emph{fine-grained spatial-temporal perception}, \emph{structured sequential planning}, and the ability to \emph{adapt strategies through interaction feedback}~\cite{huang2023vling}. However, most existing approaches still lack effective mechanisms to accumulate transferable experience across tasks: RoboOS ~\cite{tan2025roboos}primarily focuses on LLM-based orchestration without an explicit experience abstraction module, while RoboMemory ~\cite{lei2025robomemory} stores past experiences but largely emphasizes successful trajectories, overlooking systematic reflection over failure trajectories and their impact on future planning.

To address this limitation, we propose a \textbf{Brain-Inspired Evolving Memory (BrainMem)} architecture. Inspired by cognitive neuroscience~\cite{baddeley2012working,tulving1985episodic}, BrainMem introduces an \textbf{experience agent} independent of the planner that semantically abstracts both successful and failed trajectories into structured knowledge, which is stored and retrieved to guide future planning. The framework organizes memory hierarchically with working memory, episodic memory, and semantic memory to capture temporal contexts, spatial experiences, and reusable knowledge, and employs a hierarchical updating strategy to continually refine the memory for improved planning. An overview of the BrainMem architecture is shown in Fig.~\ref{fig:memory_framework}.
%


\subsection{Brain-Inspired Hierarchical Memory System}
\label{subsec:hierarchical-memory}

BrainMem is structured as a three-tier hierarchy inspired by the complementary roles of human working, episodic, and semantic memory~\cite{tulving1985episodic,baddeley2012working}. Each memory type is tasked with capturing information at a specific abstraction scale following the hierarchy of \emph{perception-planning-reflection}. These memory modules are seamlessly integrated into the agent’s planning loop as inputs, enabling it to enhance planning performance using past observations alone and without relying on external supervision.


\subsubsection{Short-Term Context Mining}
The \textbf{Working Memory} $\mathcal{M}_\text{work}$ records the necessary short-term temporal contexts for planning, analogous to human working memory for modeling short-term information~\cite{baddeley2012working}. To provide necessary beneficial information to support agent planning, we select different ingredients in $\mathcal{M}_\text{work}$ to offer various types of information. Specifically, the contents in $\mathcal{M}_\text{work}$ at each time step $t$ are consisted of:
[\text{Action},\ \text{Feedback},\ \text{Agent state},\ \text{Location},\ \text{Timestamp}]

\begin{itemize}[leftmargin=*, itemsep=0pt, topsep=1pt]
    \item \texttt{Action}, such as `pick up the book',
    \item \texttt{Feedback}, success or failure to execute,
    \item \texttt{Agent state}, holding something or holding nothing,
    \item \texttt{Location} , e.g., bedroom.
    \item \texttt{Timestamp} is the current timestep $t$ .
\end{itemize}

The above contents are combined in a sequential manner in textual formats and will be fed into the planner for planning. By preserving the most recent interaction window, the agent could keep well-acknowledged of recent observations and environmental states, which form strong support for making decisions by perceiving multi-source information.

\subsubsection{Long-Horizon Temporal Reasoning}
To capture critical long-term temporal contexts, we introduce an Episodic Memory $\mathcal{M}_\text{epi}$ that performs long-horizon reasoning with two complementary knowledge graphs (KGs).

\textbf{Trajectory KG.} It aims to accumulate experiences by observing trajectories of previously completed tasks to derive meaningful patterns across tasks. These contents are encoded as a directed graph linking consecutive action transitions between different states: [\text{Instruction},\ \text{Action sequence},\ \text{Success},\ \text{Room},\ \text{Timestamp}]

Here, \texttt{Instruction} is the textual instructions for each task. \texttt{Action sequence} is the accumulated sequence of all actions for a completed task. \texttt{Success} denotes whether the task is executed successfully or not. \texttt{Room} is the id of room when executing the task. \texttt{Timestamp} represents the concrete timestep.
These contents store the necessary information of previously completed tasks. These allow the planner to recall beneficial task-specific interactions from past experiences and derive problem-solving patterns to help resolve the current task.

\textbf{Spatial KG.}To capture persistent environment knowledge, we maintain a Spatial KG to encode the object–object and object–room relational structures of rooms visited. This allows us to retrieve structural information from our spatial KG when starting a new planning process to fully leverage past observations. Specifically, the spatial KG is constructed by recording the following two relations:
\[
(\text{room} \xrightarrow{\text{contains}} \text{object}_i), \quad
(\text{object}_i \xrightarrow{\text{near/on/in}} \text{object}_j)
\]
This graph starts empty and grows over time by recording the contents of each episode, enabling spatial reasoning across tasks.

\subsubsection{Past Experience Abstraction}

We introduce a \textbf{Semantic Memory} $\mathcal{M}_\text{sem}$ which abstracts the experiences of previous tasks into highly compact reusable guidelines, powered by LLM-based summarization.
Specifically, after each episode completes, we condense the successful pattern/failure reason of this episode into $\mathcal{M}_\text{sem}$ to enable subsequent planning processes to gain beneficial experiences from past executions.
For successful cases, we extract action sequences and key decision points that led to task completion.
These patterns are then distilled into generalizable knowledge that captures effective planning strategies applicable across similar tasks. 
For failure episodes, we employ a guided-retry mechanism that allows the agent to reattempt the task based on real-time feedback until success or a predefined retry limit is reached, capturing diverse action-outcome patterns to enrich $\mathcal{M}_\text{sem}$ with informative semantic guidelines. Importantly, task success is evaluated solely based on the first attempt, and retry outcomes do not alter the episode's failure status, ensuring fair evaluation while maximizing learning value.
Through this design, $\mathcal{M}_\text{sem}$ accumulates generalizable knowledge across tasks, mirroring memory consolidation processes in human cognition.


\subsection{Hierarchical Retrieval from Memory}

For a task query $Q=\{\text{room}, \text{task\_type}, \text{object}\}$, BrainMem retrieves relevant knowledge from multiple memory levels:

\begin{itemize}[leftmargin=*, itemsep=0pt, topsep=1pt]
    \item \textbf{Working Memory}: retain \textit{all} observations to provide closely related information for planning.
    \item \textbf{Trajectory KG}: \textit{only} recall sequences of specific past tasks with high similarity,
    \item \textbf{Spatial KG}: retrieve object-object and object-room relational records located in the \textit{same} room,
    \item \textbf{Semantic Memory}: extract prior knowledge from recorded patterns \textit{by} task type and object category,
\end{itemize}

Retrieved components are serialized into either free-form texual instructions and  appended to the LLM planner’s prompt to guide the planning procedure. This orchestrated retrieval enables memory-guided planning that fully leverage past observations and learn from past experiences.

\subsection{Feedback-Based Dyanmic Evolving}

BrainMem continually evolves the memory content in both step-level and episode-level updates, supporting the robot's ability to react promptly to environmental changes and learn from past experiences in previous episodes.

\textbf{Step-Level Updates.}
After every action completes, BrainMem updates the following contents in turn:
\begin{itemize}[leftmargin=*, itemsep=0pt, topsep=5pt]
    \item Append the interaction to Working Memory,
    \item Add resulting transitions to Trajectory KG,
    \item Update Spatial KG if new objects or relations emerge.
\end{itemize}

\textbf{Episode-Level Updates.}
After every episode completes, BrainMem conducts the following procedures in turn:
\begin{enumerate}[leftmargin=*, itemsep=0pt, topsep=5pt]
    \item Finalize Trajectory KG with the final task outcome,
    \item Extract persistent relations into Spatial KG,
    \item Summarize experiences into Semantic Memory via LLM,
    \item Add/update semantic guidelines according to usage.
\end{enumerate}

Especially, to keep useful knowledge in $M_{sem}$ and eliminate out-of-data experience, we maintain a lightweight utility score for each guideline pattern. This utility score decays over time to improve reliability on recent observations and prevent overfitting to outdated behavior. Low-utility guidelines are automatically pruned during the updating process, ensuring $\mathcal{M}_\text{sem}$ remains both compact and effective.

\section{Experiments}
\label{sec:experiments}
\begin{table*}[t]
     
    \centering\large

    \caption{
    Task success rates on 6 subsets of EB-ALFRED and EB-Habitat.  The \textbf{bold} value indicates the best value per column and the \underline{underline} value represents the best performance of previous methods.
    }
    
    \vspace{-0.2cm}
    \renewcommand{\arraystretch}{1.0}
    \setlength\tabcolsep{2pt}
    \setlength\extrarowheight{2pt}
    \resizebox{\linewidth}{!}{

    \begin{tabular}{
    >{\centering\arraybackslash}p{4cm} 
        >{\centering\arraybackslash}p{1.17cm} 
        >{\centering\arraybackslash}p{1.17cm} 
        >{\centering\arraybackslash}p{1.33cm} 
        >{\centering\arraybackslash}p{1.33cm} 
        >{\centering\arraybackslash}p{1.33cm} 
        >{\centering\arraybackslash}p{1.33cm} 
        >{\centering\arraybackslash}p{1.17cm} 
        @{\hskip 10pt} 
        >{\centering\arraybackslash}p{1.17cm} 
        >{\centering\arraybackslash}p{1.17cm} 
        >{\centering\arraybackslash}p{1.33cm} 
        >{\centering\arraybackslash}p{1.33cm} 
        >{\centering\arraybackslash}p{1.33cm} 
        >{\centering\arraybackslash}p{1.33cm} 
        >{\centering\arraybackslash}p{1.17cm} 
 }
    
        \toprule
        
         \multirow{2}{*}{\textbf{ Model}} 
         & \multicolumn{7}{c}{\textbf{EB-ALFRED}} 
         & \multicolumn{7}{c}{\textbf{EB-Habitat}} \\

        \cmidrule(lr){2-8} \cmidrule(lr){9-15}
        
        & { \textbf{Avg}} & \textbf{Base} & \textbf{Comm.} & \textbf{Comp.} & \textbf{Visual} & \textbf{Spatial} & \textbf{Long} 
        & \textbf{Avg} & \textbf{Base} & \textbf{Comm.} & \textbf{Comp.} & \textbf{Visual} & \textbf{Spatial} & \textbf{Long} \\

        \addlinespace[2pt]
        \midrule
        \addlinespace[2pt]
        \multicolumn{15}{c}{ \textit{Proprietary MLLMs} }  \\ \midrule
        {\selectfont GPT-4o} &  56.3 & 64.0 & 54.0 & 68.0 & 46.0 & 52.0 & 54.0 &   59.0  &  86.0 & 44.0 & 56.0 &68.0 &36.0 &  \underline{64.0} \\
        {\selectfont GPT-4o-mini} &  24.0 & 34.0 & 28.0 & 36.0 & 24.0 & 22.0 & 0.0 &  32.7 & 74.0 & 22.0 & 32.0  & 22.0   & 32.0 & 14.0 \\
         {\selectfont Claude-3.7-Sonnet} &  \underline{67.7} & 68.0  & \underline{68.0}  &  70.0 & \textbf{68.0}  & \underline{62.0} & \underline{70.0} &   58.7 & 90.0 & 58.0 & 58.0 &  62.0 & \underline{38.0} & 46.0  \\
        {\selectfont Claude-3.5-Sonnet} &  64.0 & \underline{72.0}  & 66.0 &  \underline{76.0} & 60.0  & 58.0 & 52.0 &   \underline{68.0} & \textbf{96.0} & \underline{68.0} &  \underline{78.0} & \underline{70.0} & \underline{38.0} & 58.0 \\
        {\selectfont Gemini-1.5-Pro} &  62.3 & 70.0  & 64.0  & 72.0 &  58.0  &  52.0 & 58.0 &   56.3 & 92.0 & 52.0 & 48.0 & 56.0 &  \underline{38.0} &  52.0   \\
        {\selectfont Gemini-1.5-flash} &  39.3 & 44.0  & 40.0 & 56.0 & 42.0 & 26.0 & 28.0 &  39.3 & 76.0 & 32.0 & 48.0 & 36.0 & 32.0 & 12.0 \\

        {\selectfont Qwen-VL-Max } &  41.3 &  44.0 & 48.0 & 44.0 & 42.0 & 38.0 &  32.0 &  45.3 & 74.0 & 40.0  & 50.0 & 42.0 & 30.0 & 36.0 \\
       
        \midrule
        \addlinespace[2pt]
        
        \multicolumn{15}{c}{ \textit{Open-Source MLLMs} }   \\ \midrule
         {\selectfont Llama-3.2-90B-Vision} &  32.0 & 38.0 & 34.0 & 44.0 & 28.0 &  32.0 & 16.0 &  40.3  &   94.0 & 24.0 & 50.0 & 32.0 & 28.0 & 14.0 \\
         
       {\selectfont InternVL2\_5-78B} &  37.7 & 38.0  & 34.0 & 42.0  & 34.0  & 36.0 & 42.0 &  49.0 & 80.0 & 42.0 &  56.0 & 58.0 & 30.0 & 28.0 \\

{\selectfont InternVL3-78B } &  39.0 & 38.0 & 34.0 & 46.0 & 42.0 & 38.0 & 36.0 &   55.0 & 84.0 & 58.0 & 60.0 & 56.0 & 32.0 & 40.0  \\

{\selectfont Qwen2.5-VL-72B} &  39.7 & 50.0 & 42.0 & 42.0 & 36.0 & 34.0 & 34.0&  37.7 & 74.0 & 28.0 & 42.0 & 40.0 & 24.0 & 18.0\\

{\selectfont Ovis2-34B } &  28.7 & 34.0 & 30.0 & 38.0 & 28.0 & 18.0 & 24.0 &  37.0 & 68.0 & 34.0 & 38.0 &  38.0 & 30.0 & 14.0 \\


\midrule
    \addlinespace[2pt]
    \multicolumn{15}{c}{ \textit{Proprietary MLLMs with BrainMem} } \\
    \midrule

    {\selectfont GPT-4o}
    &  75.0 & 82.0 & 78.0 & 82.0 & 64.0 & \textbf{72.0} & 72.0
    &  63.3 & 90.0 & 46.0 & 60.0 & \textbf{74.0} & 42.0 & \textbf{68.0}\\

    {\selectfont Claude-3.5-Sonnet}
    &  74.7 & 78.0 & 72.0 & \textbf{84.0} & 66.0 & 64.0 & \textbf{84.0}
    &  \textbf{73.0} & \textbf{96.0} & \textbf{74.0}& \textbf{84.0} & \textbf{74.0} & \textbf{46.0} & 64.0 \\

    {\selectfont Gemini-1.5-pro}
    &  \textbf{75.3} & \textbf{88.0} & \textbf{80.0} & 80.0 & \textbf{68.0} & \textbf{72.0} & 64.0
    &  64.3 & 94.0 & 60.0 & 58.0 & 68.0 & \textbf{46.0} & 60.0\\

        \bottomrule
        
    \end{tabular}
    }\label{tb:high_level_table}
    
    \vspace{-20pt}

\end{table*}

\begin{table*}[t]

    \centering\large
    \caption{
    Task success rates on 5 subsets of EB-Navigation and EB-Manipulation. The \textbf{bold} value indicates the best value per column and the \underline{underline} value represents the best performance of previous methods.
    }
    \vspace{-10pt}
    \renewcommand{\arraystretch}{1.1}
    \setlength\tabcolsep{2pt}
    \setlength\extrarowheight{2pt}
    \resizebox{\linewidth}{!}{

        \begin{tabular}{
    >{\centering\arraybackslash}p{4cm} 
        >{\centering\arraybackslash}p{1.17cm} 
        >{\centering\arraybackslash}p{1.17cm} 
        >{\centering\arraybackslash}p{1.33cm} 
        >{\centering\arraybackslash}p{1.33cm} 
        >{\centering\arraybackslash}p{1.33cm} 
        >{\centering\arraybackslash}p{1.33cm} 
        >{\centering\arraybackslash}p{1.17cm} 
        @{\hskip 10pt} 
        >{\centering\arraybackslash}p{1.17cm} 
        >{\centering\arraybackslash}p{1.33cm} 
        >{\centering\arraybackslash}p{1.33cm} 
        >{\centering\arraybackslash}p{1.33cm} 
        >{\centering\arraybackslash}p{1.33cm} 
        >{\centering\arraybackslash}p{1.17cm} 
 }
    
        \toprule
        
        \multirow{2}{*}{\textbf{Model}} 
        & \multicolumn{6}{c}{\textbf{EB-Navigation}} 
        & \multicolumn{6}{c}{\textbf{EB-Manipulation}} \\
        
        \cmidrule(lr){2-7} \cmidrule(lr){8-13}
        
        &  \textbf{Avg} 
        & \textbf{Base} 
        & \textbf{Comm.} 
        & \textbf{Comp.} 
        & \textbf{Visual}  
        & \textbf{Long} 
        &  \textbf{Avg} 
        & \textbf{Base} 
        & \textbf{Comm.} 
        & \textbf{Comp.} 
        & \textbf{Visual} 
        & \textbf{Spatial}  \\

        \addlinespace[2pt]
        \midrule
        \addlinespace[2pt]
        
        \multicolumn{13}{c}{ \textit{Proprietary MLLMs} }  \\ 
        \midrule
        
        {\selectfont GPT-4o} &  \underline{57.7} & 55.0 & 60.0 & 58.3 & \underline{60.0}  & \underline{55.0} 
        &  \underline{28.9} & \underline{39.6} & \underline{29.2} & 29.2 & 19.4 & 25.0 \\ 
        {\selectfont GPT-4o-mini} &   32.8 & 31.7 & 33.3 & 35.0 & 28.3 & 33.3 &  4.8 & 4.2 & 6.3 & 2.1 & 0.0 & 10.4 \\
        {\selectfont Claude-3.7-Sonnet} &  45.0 & 50.0 & 61.7 & 50.0 & 36.7 & 26.7 &   28.5 & 31.3 & 20.8 & \underline{43.8} & 25.0 & 20.8 \\
        {\selectfont Claude-3.5-Sonnet}&   44.7 & \underline{66.7} & 51.7 & 41.7 & 36.7 & 26.7
        &  25.4 & 37.5 & 16.7 & 29.2 & 19.4 & 22.9 \\
        {\selectfont Gemini-1.5-Pro}  &  24.3 &23.3 &25.0&25.0& 28.3&20.0
        &  21.1 & 14.6 & 14.6 & 22.9 & 16.7 & \underline{35.4} \\
        {\selectfont Gemini-1.5-flash} &  41.7 & 56.7 & 50.0 & 46.7 & 50.0 & 5.0
        &  9.6 & 14.6 & 10.4 & 4.2 & 8.3 & 10.4 \\
        {\selectfont Qwen-VL-Max } &   39.7 & 50.0 & 46.7 & 41.7 & 35.0 & 25.0 &  18.0 & 25.0 & 10.4 & 18.8 & 2.8 & 29.2 \\
       
        \midrule
        \addlinespace[2pt]
        
        \multicolumn{13}{c}{ \textit{Open-Source MLLMs} }   
        \\ \midrule
        

         {\selectfont Llama-3.2-90B-Vision} &  30.0 & 48.3 & 23.3 & 38.3 & 33.3 & 6.7 
        &  14.9 & 10.4 & 12.5 & 16.7 & 10.4 & 20.8 \\
        
        {\selectfont InternVL2\_5-78B} &  30.7 & 36.7 & 38.3 & 33.3 & 21.7 & 23.3 
        &  18.0 & 16.7 & 16.7 & 14.6 & 22.2 & 20.8 \\

        {\selectfont InternVL3-78B} &  53.7 & \underline{66.7} & 63.3 & \underline{61.7} & 45.0 & 31.7 
        &  26.3 & 29.2 & 22.9 & 22.9 & 25.0 & 31.3 \\

        {\selectfont Qwen2.5-VL-72B} &  40.0 & 46.7  & 46.7 & 46.7 & 26.7 & 33.3 &  16.2 & 12.5 & 12.5 & 16.7 & 22.2 & 18.8 \\
        
        {\selectfont  Ovis2-34B} &  45.7  & 63.3 & 50.0 & 56.7 & 46.7 & 11.7 &  26.8 & 31.3 & 25.0 & 18.8 & \underline{27.8} & 31.3 \\
       
        
        \midrule
        \addlinespace[2pt]
        \multicolumn{13}{c}{ \textit{Proprietary MLLMs with BrainMem} } \\
        \midrule
    
        {\selectfont GPT-4o}
        &  \textbf{71.0} & \textbf{73.3} &  71.7 & \textbf{80.0} & 66.6 & \textbf{63.3}
        &  \textbf{46.9} & \textbf{47.9} & \textbf{50.0} & \textbf{47.9}  & \textbf{47.2} & 47.9  \\
    
        {\selectfont Claude-3.5-Sonnet}
        &  63.3 & 70.0 & \textbf{73.3} & 73.3 & \textbf{68.3} & 31.7
        &  43.1 & 45.8 & 43.8 & 37.5 & 44.4 & 43.8 \\
    
        {\selectfont Gemini-1.5-pro}
        &  61.0 & 71.7 & 68.3 & 68.3 & 51.7 & 45.0
        &  38.6 & 33.3 & 39.6 & 35.4 & 30.6 & \textbf{54.2} \\

        \bottomrule
    \end{tabular}\label{tb:low_level_table}
    }
    
   \vspace{-25pt}

\end{table*}

\subsection{Experimental Setup}
\label{ssec:setup}
\textbf{Environments and Datasets.}
We evaluate our approach across four representative embodied benchmarks including \textbf{EB-ALFRED}, \textbf{EB-Habitat}, \textbf{EB-Navigation}, and\textbf{EB-Manipulation} , which are built upon the AI2-THOR simulator~\cite{kolve2017ai2}.  
Specifically,
EB-ALFRED consists of long-horizon household task-planning tasks. EB-Habitat extends the embodied setting to large-scale realistic 3D indoor environments with continuous control and navigation. EB-Navigation focuses on multi-room goal-conditioned navigation. EB-Manipulation emphasizes fine-grained object interaction.  Each dataset is divided into multiple subsets of varying difficulty levels. EB-ALFRED and EB-Habitat each include six subsets—\textit{Base, Common, Complex, Visual, Spatial, Long}—with 50 held-out test instances. EB-Navigation comprises 300 test cases across five subsets—\textit{Base, Common, Complex, Visual, Long} (60 each), while EB-Manipulation contains 228 instances, with 48 per subset for—\textit{Base, Common, Complex, Spatial}  and 36 for subset \textit{Visual}.

\textbf{Evaluation Protocol.}
Following prior works~\cite{alfred}, we adopt Success Rate (SR) as the primary evaluation metric, which is defined as the percentage of episodes that successfully achieve both the final task goal and all required object interactions.

\textbf{Planners and Baselines.}
We evaluate the embodied task planning performance across a diverse set of proprietary and open-source multi-modal LLM planners. For proprietary models, we directly evaluate GPT-4o~\cite{gpt4o}, Claude-3.5-Sonnet~\cite{claude3.5}, and Gemini-1.5-Pro~\cite{gemini1.5}. For open-source MLLMs, we include Llama-3.2~\cite{dubey2024llama}, InternVL3~\cite{zhu2025internvl3}, Qwen2.5-VL~\cite{bai2025qwen2}, Ovis2~\cite{lu2025ovis2}, and others for comparison.
All proprietary evaluations are conducted using a unified controller API for fair comparison. For consistency, all models are set with a temperature of 0 and a maximum completion length of 2048 tokens. All experiments use standardized 500×500 visual inputs and fixed step limits (30/20/15 for high-level, navigation, and manipulation tasks, respectively) following EmbodiedBench~\cite{yang2025embodiedbench}.


\subsection{Main Results}
\label{ssec:main_results}

\textbf{Performance Gains with BrainMem.} 
Tables~\ref{tb:high_level_table} and~\ref{tb:low_level_table} summarize task success rates on EB-ALFRED / EB-Habitat and EB-Navigation / EB-Manipulation. 
The top sections report baseline results from both proprietary and open-source MLLMs for benchmarking, while the bottom sections show results of the same proprietary models augmented with our BrainMem memory architecture. BrainMem consistently improves success rates across state-of-the-art planners. On \textbf{EB-ALFRED}, performance increases to around \textbf{75\%} across GPT-4o, Claude-3.5-Sonnet, and Gemini-1.5-Pro. On \textbf{EB-Habitat}, it brings stable gains of \textbf{+4–8\%}. Even larger improvements appear on more challenging tasks, reaching up to \textbf{+36.7\%} on \textbf{EB-Navigation} and about \textbf{+17–18\%} on \textbf{EB-Manipulation}.

\textbf{Comparison with previous memory-related methods.}  
Table~\ref{tab:combined_full_benchmarks} compares BrainMem with previous memory-related methods using the same backbone, \textit{Qwen2.5-VL-72B-Ins}, and reports their performance on four subsets (EB-ALFRED \textit{Base/Long} and EB-Habitat \textit{Base/Long}). BrainMem achieves the best overall performance (\textbf{75.0\% SR / 82.2\% GC}) among memory-based methods. On EB-ALFRED, the improvements are more pronounced on the challenging \textit{Long} subset, while also providing consistent gains on \textit{Base} tasks. On EB-Habitat, although the \textit{Long} subset shows same SR to RoboMemory, BrainMem achieves a higher GC, indicating more reliable progress toward task completion.
Compared with prior methods, BrainMem introduces a hierarchical memory architecture that integrates \textit{working}, \textit{episodic}, and \textit{semantic} memory, together with an independent experience agent that abstracts both successful and failed trajectories into reusable knowledge. Unlike Reflexion or Voyager, which rely mainly on textual feedback loops, RoboMemory stores past experiences but largely focuses on successful trajectories, leaving failure reflection insufficiently abstracted for reuse. In contrast, BrainMem enables structured experience accumulation and hierarchical retrieval, improving long-term reasoning and spatial-temporal understanding.

\subsection{Ablation Studies}
\label{ssec:ablation}

Unless otherwise stated, we conduct ablations with GPT-4o on the base set of EB-ALFRED benchmark to evaluate the effectiveness of each component.

\begin{table}[t]
\caption{Comparison of Success Rates (SR) and Goal Condition Success Rates (GC) across difficulty levels (Base/Long) on EB-ALFRED and EB-Habitat. All methods are evaluated with Qwen2.5-VL-72B-Ins. Values are reported in percentages (\%).}
\label{tab:combined_full_benchmarks}

\vspace{-5pt}
\centering
\setlength{\tabcolsep}{2.5pt} 
\begin{tabular}{l|cc|cccc|cccc}
\toprule
\multirow{3}{*}{Method} &
\multicolumn{2}{c}{} & 
\multicolumn{4}{c}{EB-ALFRED} & 
\multicolumn{4}{c}{EB-Habitat} \\
\cmidrule(lr){4-7} \cmidrule(lr){8-11} 

&  
\multicolumn{2}{c}{Average} & 
\multicolumn{2}{c}{Base} & \multicolumn{2}{c}{Long} & 
\multicolumn{2}{c}{Base} & \multicolumn{2}{c}{Long} \\
\cmidrule(lr){2-3} \cmidrule(lr){4-5} \cmidrule(lr){6-7} \cmidrule(lr){8-9} \cmidrule(lr){10-11}

& SR & GC & SR & GC & SR & GC & SR & GC & SR & GC \\
\midrule


Voyager (NeurIPS 2023) & 46.5 & 66.4 & 56.0 & 73.2 & 32.0 & 54.2 & 76.0 & 87.0 & 22.0 & 51.0 \\
Reflexion (NeurIPS 2023)  & 38.3 & 51.1 & 48.0 & 54.0 & 10.0 & 33.0 & 80.0 & 84.2 & 15.0 & 33.0 \\
Cradle (ICLR 2024) & 44.5 & 57.0 & 54.0 & 67.9 & 32.0 & 41.0 & 62.0 & 67.0 & 30.0 & 52.1 \\
RoboOS (2025.6) & 25.5 & 33.0 & 32.0 & 38.4 & 12.0 & 17.6 & 38.0 & 47.8 & 20.0 & 28.2 \\
RoboMemory (2025.10) & \underline{70.5} & \underline{79.7} & \underline{68.0} & \underline{75.5} & \underline{66.0} & \underline{81.3} & \underline{86.0} & \underline{88.0} & \textbf{62.0} & \underline{74.0} \\
\midrule
\textbf{BrainMem (Ours)} & \textbf{75.0} & \textbf{82.2} & \textbf{72.0} & \textbf{78.2} & \textbf{74.0} & \textbf{82.6} & \textbf{92.0} & \textbf{92.6} & \textbf{62.0} & \textbf{75.3}  \\
\bottomrule
\end{tabular}
\vspace{-2em}
\end{table}

\textbf{Effect of each memory module.}
We first validate the effectiveness of our proposed memory modules by progressively adding each one to build the final model.  As shown in Table~\ref{tb:ablation_memory}, stepwisely adding the working memory, episodic \begin{wraptable}{r}{0.48\textwidth}
\vspace{-34pt}
\centering
\caption{Ablation of memory modules on EB-ALFRED. Success rates (\%) are averaged over six subsets.}
\vspace{-0.cm}
\footnotesize
\renewcommand{\arraystretch}{1.1}
\resizebox{0.48\textwidth}{!}{
\begin{tabular}{ccccc}
\toprule
\makecell[c]{Working \\ Memory} & 
\makecell[c]{Episodic \\ memory} & 
\makecell[c]{Semantic \\ memory} & 
Feedback & 
\makecell[c]{SR (\%)} \\
\midrule
\ding{55} & \ding{55} & \ding{55} & \ding{55} & 64.0 \\
\ding{51} & \ding{55} & \ding{55} & \ding{55} & 68.0 \\
\ding{51} & \ding{51} & \ding{55} & \ding{55} & 70.0 \\
\ding{51} & \ding{51} & \ding{51} & \ding{55} & 76.0 \\
\ding{51} & \ding{51} & \ding{51} & \ding{51} & \textbf{82.0} \\
\bottomrule
\end{tabular}}
\label{tb:ablation_memory}
\vspace{-30pt}
\end{wraptable} memory, and semantic memory consistently yields notable performance gains by boosting the performance from 64.0\% (no memory) to 76.0\% (+12.0\%), which strongly prove the  effectiveness of each proposed  module. Moreover, we observe adding the feedback mechanism brings extra performance boost (+6.0\%),
which verifies that dynamically evolving memory contents could allow the model to observe the environments to make reliable decisions throughly.

\textbf{Analysis for the contents in working memory.}
We validate the effects of each component in the working memory in
Table~\ref{tb:ablation_wm_content}  to testify their effectiveness. It shows that removing any of the components, including the feedback, agent state, success rate, and consistency check, consistently leads to lower success rates, which indicate that each component brings unique and crucial information for the robot to capture short-term contexts to make coherent action planning.

\textbf{Ablations for the window size of working memory.}
We ablate the effect of the working memory window size on performance. As shown in Table~\ref{tb:ablation_window}, performance improves as the window increases from 1 and peaks at 5. Although sizes 3 and 5 achieve identical overall success rates, size 5 yields a higher Subgoal Success Rate (84.3\% vs. 82.6\%), indicating better intermediate progress. Larger windows degrade performance, likely due to redundant contextual information. We therefore set the default window size to 5.



\begin{table}[t]


\newcommand{\subtblw}{0.43\linewidth} 
\begin{minipage}[t]{\subtblw}
\centering
\captionof{table}{Ablation for the contents in working memory.}
\vspace{-0.2cm}
\footnotesize
    \centering
    \setlength\tabcolsep{5pt}
\resizebox{0.95\textwidth}{!}{
    \begin{tabular}{l c}
        \toprule
        Configuration & SR (\%) \\
        \midrule
        Final model        & \textbf{82.0} \\
        \ - w/o feedback               & 76.0 \\
        \ - w/o success state          & 80.0   \\
        \ - w/o agent state            & 74.0   \\
        \ - w/o consistency check      & 70.0   \\
        \bottomrule
    \end{tabular}
    }
    \label{tb:ablation_wm_content}
\end{minipage}
\hfill
\begin{minipage}[t]{0.55\linewidth}
\centering
\captionof{table}{Ablation for the window size of working memory.}
\vspace{-0.2cm}
\footnotesize
\resizebox{0.91\textwidth}{!}{
    \begin{tabular}{ccc}
        \toprule
        Window Size & SR (\%) & Goal Condition SR(\%)\\
        \midrule
        1 & 72.0 & 73.3 \\
        3 & \textbf{82.0} & 82.6\\
        5 & \textbf{82.0} & \textbf{84.3} \\
        8 & 76.0 & 76.7 \\
        10 & 78.0 & 80.2\\
        \bottomrule
    \end{tabular}
}
    \label{tb:ablation_window}
\end{minipage}
\vspace{-1em}
\end{table}


\textbf{Effects of different knowledge graphs in episodic memory.} Different knowledge graphs in episodic memory play different roles in providing episodic-level information for the robot. The trajectory knowledge graph offers cross-episodic experiences containing various ingredients while the spatial knowledge records spatial relationships between objects and rooms. As shown in
Table~\ref{tb:ablation_em_kg}, removing each knowledge graph leads to considerable performance drop, which underscore their complementariness and effectiveness in grounding action reasoning and environment understanding.


\begin{table}[t]
\centering
\newcommand{\subsmall}{0.3\linewidth}
\newcommand{\sublarge}{0.23\linewidth}
\newcommand{\sublargee}{0.44\linewidth}

\begin{minipage}[t]{\subsmall}
\centering
\captionof{table}{The effect of disabling different types of knowledge graph (KG) in episodic memory.}
\vspace{-0.2cm}
\scriptsize
\renewcommand{\arraystretch}{1.31}
\begin{tabular}{l c}
    \toprule
    Configurations & SR (\%) \\
    \midrule
    Final model       & \textbf{82.0} \\
    \ - w/o spatial KG           & 80.0   \\
    \ - w/o trajectory KG & 78.0   \\
    \bottomrule
\end{tabular}
\label{tb:ablation_em_kg}
\end{minipage}
\hfill
\begin{minipage}[t]{\sublarge}
\centering
\captionof{table}{The effects of retrieving top-$k$ experience in semantic memory.}
\vspace{-0.2cm}
\scriptsize
\renewcommand{\arraystretch}{1.05}
\setlength{\tabcolsep}{4pt}
\begin{tabular}{c c}
    \toprule
    Top-$k$ & SR (\%) \\
    \midrule
    1 & 74.0\\
    2 & \textbf{82.0}\\
    3 & \textbf{82.0}\\
    5 & \textbf{82.0}\\
    \bottomrule
\end{tabular}
\label{tb:ablation_topk}
\end{minipage}
\hfill
\begin{minipage}[t]{\sublargee}
\centering
\captionof{table}{Ablation study on the effects of enabling different combinations of components in semantic memory during task execution}
\vspace{-0.2cm}
\scriptsize
\setlength{\tabcolsep}{4pt}
\renewcommand{\arraystretch}{0.645}
\begin{tabular}{cccc}
    \toprule
    \makecell[c]{Successful \\ Patterns} & 
    \makecell[c]{Failure \\ Reasons} & 
    \makecell[c]{Learning \\ Insights} & 
    \makecell[c]{SR (\%)} \\
    \midrule
    \ding{51} & \ding{51} & \ding{51} & 82.0 \\
    \ding{55} & \ding{51} & \ding{51} & 80.0 \\
    \ding{51} & \ding{55} & \ding{51} & 74.0 \\
    \ding{51} & \ding{51} & \ding{55} & 72.0 \\
    \bottomrule
\end{tabular}
\label{tb:ablation_experience}
\end{minipage}
\vspace{-0.8cm}
\end{table}


\textbf{Analysis for the top-$k$ depth of retrieving experiences.}
Table~\ref{tb:ablation_topk} ablates the effects of retrieving various top-$k$ experiences from the semantic memory. We find that the performance continue increasing as the retrieval depth grows from 1, and retrieving two most relevant past experiences ($k{=}2$) provides the best trade-off between diversity and conciseness. We thus adopt $k{=}2$ by default.

\textbf{Ablations for the effectiveness of components in semantic memory.} Tab.~\ref{tb:ablation_experience} removes each component including the successful pattern, failure reasons and learning insight from semantic memory to validate their effectiveness. We observe that removing each would lead to a considerable performance drop, and disabling the learning insights decreases the performance the most. This validates that enabling the robot agent to consolidate useful guidelines from past experiences could notably improve target task performance. 

\textbf{Efficiency and latency.} 
BrainMem incurs $\sim$1.4$\times$ token \begin{wraptable}{r}{0.48\textwidth}
\vspace{-34pt}
    \centering
\captionof{table}{Total LLM token usage (input + output) on EB-ALFRED.}
\footnotesize
    \centering
    \resizebox{0.5\textwidth}{!}{
\begin{tabular}{lcc}
\toprule
Method & \textit{Base} subset & \textit{Long} subset\\
\midrule
Baseline (no memory) & 2.08 $\times 10^6$ & 4.20$\times 10^6$ \\
\textbf{+BrainMem} & 2.92$\times 10^6$ & 5.89$\times 10^6$\\
\bottomrule
\end{tabular}
}
    \label{tb:rebuttal_tokens}
\vspace{-20pt}
\end{wraptable} overhead (Table~\ref{tb:rebuttal_tokens}). Memory operations add negligible latency via lightweight symbolic manipulation.  Furthermore, LLM summarization occurs post-episode, ensuring zero impact on real-time planning latency.


\subsection{Case Study and Visualization}
\label{ssec:visualization}

\begin{figure*}[t]
    \centering
    \includegraphics[width=1\linewidth]{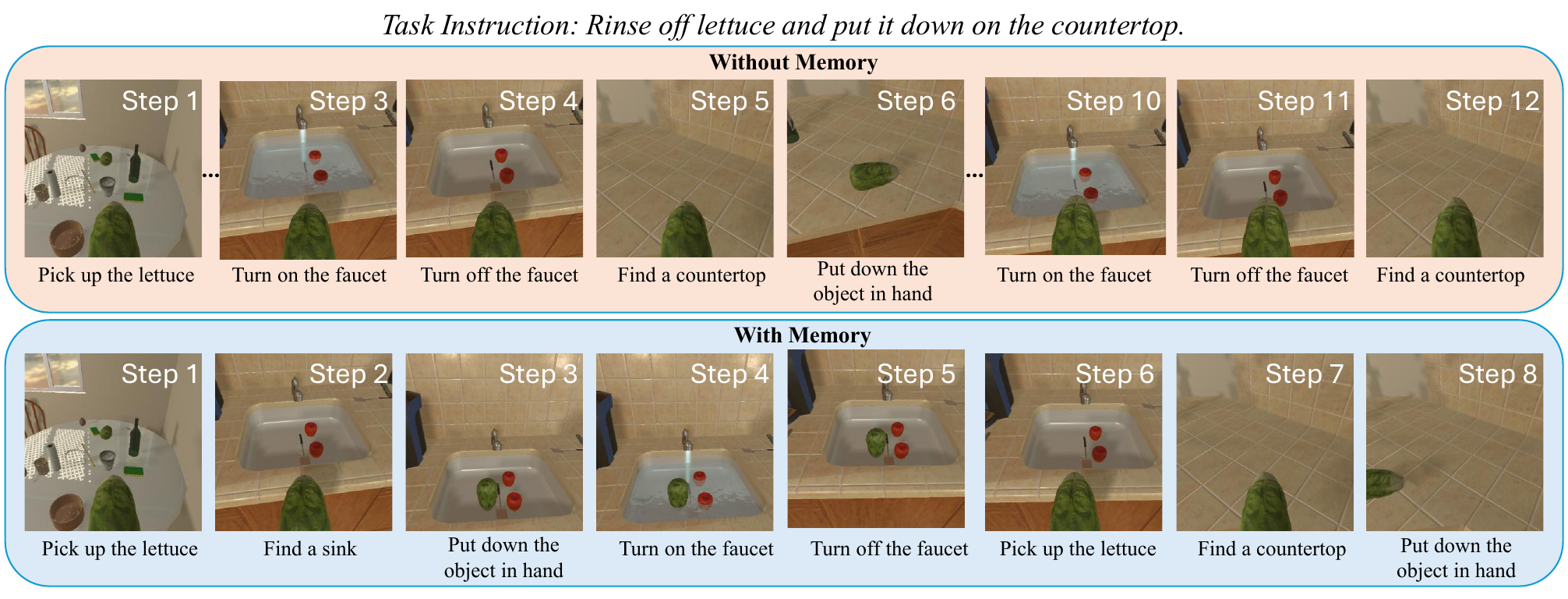}
    \vspace{-0.5cm}
    \caption{\textbf{Visualization of task execution without and with memory.} 
    The example task is \textit{“Rinse off lettuce and put it down on the countertop.”}
    Without memory (top), the agent repeatedly opens and closes the faucet without placing the lettuce inside, failing to complete the task.
    With memory (bottom), the agent recalls that the object must be placed in the sink before turning on the faucet, executing the correct sequence.}
    \label{fig:visualization}
    \vspace{-2em}
\end{figure*}

To better understand how hierarchical memory contributes to adaptive reasoning, we visualize the complete execution trace for the task \textit{“Place the lettuce in the sink.”} Figure~\ref{fig:visualization} compares trajectories with and without BrainMem memory.

Without memory (top), the agent loses contextual awareness and repeats incorrect actions. It turns on the faucet before placing the lettuce in the sink, mistakenly assumes the task is finished after leaving the lettuce on a counter, and falls into an endless cycle of redundant steps. This behavior typifies stateless planners that cannot track short-term progress or correct local errors.

With memory (bottom), the agent recalls prior experiences about cleaning and soaking objects. Through the coordination of working and episodic memory, it verifies that the lettuce must first be in the sink before turning on the faucet. The agent then executes the sequence correctly and completes the task successfully. This qualitative comparison illustrates how BrainMem grounds reasoning in both the recent context and the accumulated experience.

\subsection{Error Analysis}
\label{ssec:error}
\begin{figure}[t]
    \centering
    \includegraphics[width=1\linewidth]{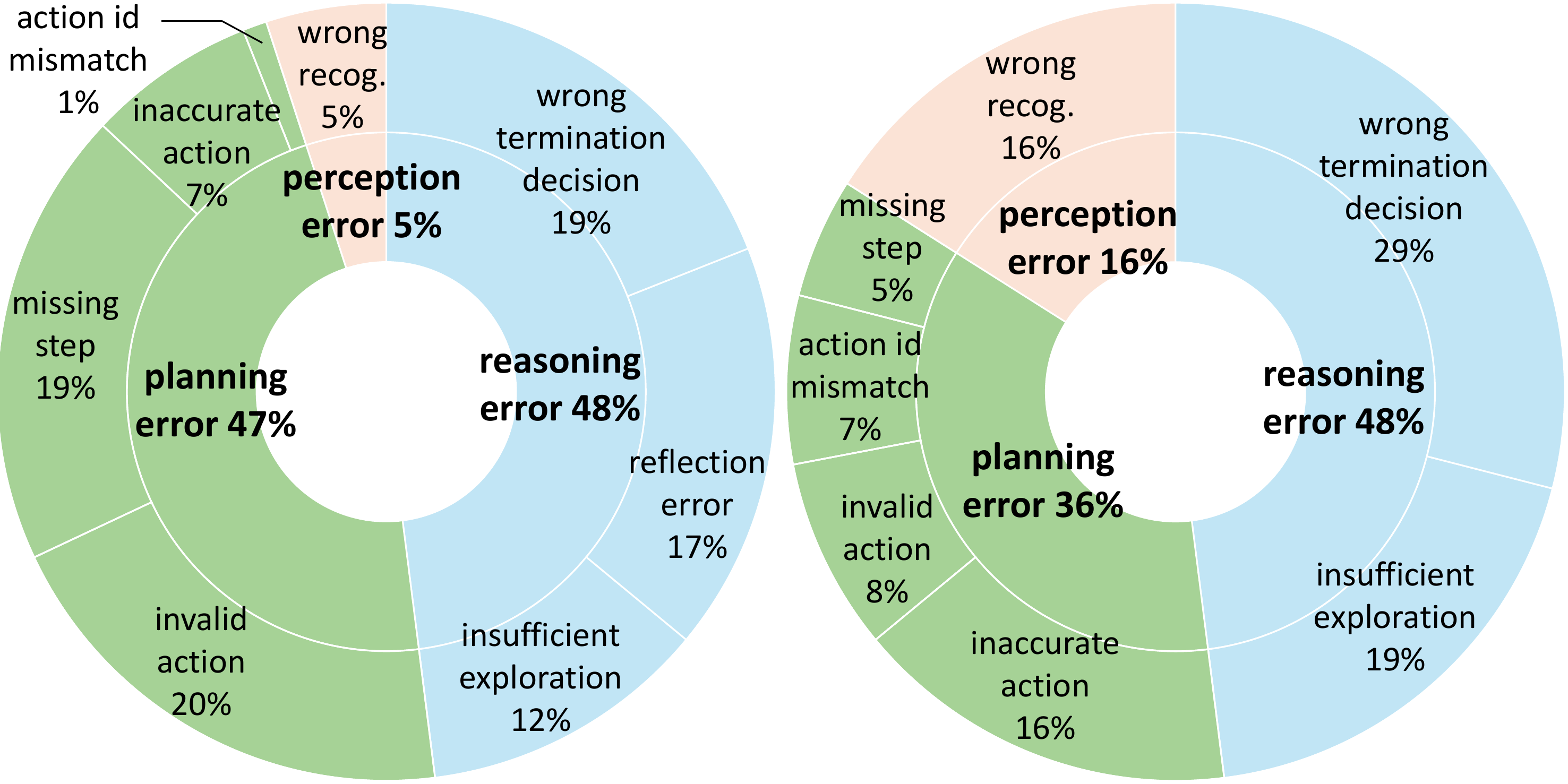}
    \vspace{-0.3cm}
    \caption{\textbf{Error distribution on EB-ALFRED using GPT-4o.} 
    The left chart shows the breakdown of 131 failure cases without memory, dominated by reasoning and planning errors. 
    The right chart shows 75 failure cases with BrainMem memory, where planning errors are notably reduced and reflection errors disappear, demonstrating the stabilizing effect of hierarchical memory.}
    \label{fig:error}
    \vspace{-0.6cm}
\end{figure}

To further understand how hierarchical memory improves task robustness, we conduct a fine-grained error analysis on failed episodes using GPT-4o as the planner in EB-ALFRED. We categorize failures into three types: \textbf{perception}, \textbf{reasoning}, and \textbf{planning}, and annotate subtypes such as misrecognition, invalid actions, and missing steps. Figure~\ref{fig:error} compares error distributions for agents without and with BrainMem, based on 131 and 75 failures, respectively.

Without memory, reasoning and planning errors dominate. The agent often misjudges task termination (19\%) or executes invalid or missing actions (20\% and 19\%), reflecting the instability of stateless decision-making. The absence of working memory also leads to repetitive behaviors, while perception errors (5\%) occasionally trigger cascading reasoning failures.

With BrainMem, reasoning and planning errors decrease substantially, and total failures drop from 131 to 75. Remaining errors shift toward higher-level reasoning, particularly incorrect termination (29\%) and mild perception mismatches (16\%), suggesting that BrainMem mitigates procedural and action-level faults but still struggles with long-horizon goal inference. Notably, reflection-related errors vanish entirely, indicating improved self-consistency through memory-guided reasoning. Overall, hierarchical memory stabilizes low-level planning while revealing room for improving abstract reasoning.

\section{Conclusion}
We presented BrainMem (Brain-Inspired Evolving Memory), a hierarchical memory architecture that equips embodied agents with the ability to remember, reason, and adapt like humans. BrainMem introduces working, episodic, and semantic memory modules that collectively support short-term context maintenance, long-term experience accumulation, and abstract knowledge formation without additional training or architectural modification. Through continuous memory evolution and retrieval, BrainMem enables embodied planners to refine strategies, recover from errors, and generalize behaviors across diverse tasks and environments. Extensive experiments across EB-ALFRED, EB-Habitat, EB-Navigation, and EB-Manipulation demonstrate that BrainMem consistently improves task success rates over stateless baselines, especially on long-horizon and spatially complex scenarios. Qualitative analyses further reveal that BrainMem enhances coherence, reduces repetitive failures, and fosters self-corrective behavior by grounding decision-making in structured experiential knowledge.

%
%
\bibliographystyle{splncs04}
\bibliography{main}
\appendix
\newpage
\setcounter{page}{1}

\begin{center}
    {\textbf{Supplementary Materials for ``BrainMem: Brain-Inspired Evolving Memory for Embodied Agent Task Planning''}}
\end{center}

This supplementary document provides additional experimental results, visualizations, and implementation details complementing the main paper. It is organized as follows:
\begin{itemize}
    \item Section A: Additional Experimental Results  
    \item Section B: Extended Method Details  
    \item Section C: Prompt Templates and Examples  
    \item Section D: Additional Visualizations  
\end{itemize}

\section*{A. Additional Experimental Results}

\subsection*{A.1 Subgoal Success Rate}

Table~\ref{tb:suppl_1_subgoal} reports the subgoal success rates (\%) on the six EB-ALFRED subsets and the six EB-Habitat subsets.  
We include both proprietary and open-source multimodal LLMs, showing performance with and without BrainMem where applicable.

Subgoal success rate provides a finer-grained measure than final task success, as it reflects the agent's ability to execute intermediate steps such as picking, placing, cleaning, heating, and navigating between rooms. This metric reveals how memory affects procedural reliability even when the final task may fail.

Our supplementary analysis includes:
\begin{itemize}
    \item Complete subgoal success rates for all model families across all difficulty subsets.  
    \item A comparison between subgoal success gains and full-task success gains under BrainMem augmentation.  
    \item Observations on model-specific behaviors (e.g., improvements in cleaning, placing, and spatial reasoning tasks).  
\end{itemize}

\begin{table*}[t]
    \centering
    \small
    \label{tb:suppl_subgoal}
    \vspace{0.3em}
        \centering\small
    \caption{
    \textbf{Subgoal success rates} on 6 subsets of EB-ALFRED and EB-Habitat, with the best proprietary model in bold and open-source model underlines per column. 
    }\label{tb:suppl_1_subgoal}
    \renewcommand{\arraystretch}{1.0}
    \setlength\tabcolsep{2pt}
    \setlength\extrarowheight{2pt}
    \resizebox{\linewidth}{!}{
    \begin{tabular}{
     >{\centering\arraybackslash}p{3.3cm} 
        >{\centering\arraybackslash}p{1.17cm} 
        >{\centering\arraybackslash}p{1.17cm} 
        >{\centering\arraybackslash}p{1.33cm} 
        >{\centering\arraybackslash}p{1.33cm} 
        >{\centering\arraybackslash}p{1.33cm} 
        >{\centering\arraybackslash}p{1.17cm} 
        >{\centering\arraybackslash}p{1.17cm} 
    @{\hskip 10pt} 
      >{\centering\arraybackslash}p{1.17cm} 
        >{\centering\arraybackslash}p{1.17cm} 
        >{\centering\arraybackslash}p{1.33cm} 
        >{\centering\arraybackslash}p{1.33cm} 
        >{\centering\arraybackslash}p{1.33cm} 
        >{\centering\arraybackslash}p{1.33cm} 
        >{\centering\arraybackslash}p{1.17cm} 
 }
    
        \toprule
        
         \textbf{\small Model} & \multicolumn{7}{c}{ \bf EB-ALFRED} & \multicolumn{7}{c}{ \bf EB-Habitat} \\

        \cmidrule(lr){2-8} \cmidrule(lr){9-15}
        
        ~ & \textbf{Avg} & \textbf{Base} & \textbf{Common} & \textbf{Complex} & \textbf{Visual} & \textbf{Spatial} & \textbf{Long} 
       & \textbf{Avg} & \textbf{Base} & \textbf{Common} & \textbf{Complex} & \textbf{Visual} & \textbf{Spatial} & \textbf{Long}  \\

        \addlinespace[2pt]
        \midrule
        \addlinespace[2pt]
        \multicolumn{15}{c}{ \textit{Proprietary MLLMs} }  \\ \midrule
        {\fontsize{8}{10}\selectfont GPT-4o} & 65.1 & 74.0&  60.3 & 74.0  & 58.3 & \underline{61.3} & 62.5 & 70.7 & 90.7 &  56.0 & 68.0 & \underline{75.2} &  \underline{62.1} & \underline{72.2}  \\
        {\fontsize{8}{10}\selectfont GPT-4o-mini} & 34.3 & 47.8 & 35.3 & 43.5 & 33.3 & 29.0 &  17.0 &  44.0 &  77.5 &32.5 & 42.0 & 33.1 & 57.8 & 21.3   \\
        {\fontsize{8}{10}\selectfont Claude-3.5-Sonnet} & 65.3 & 72.0 & 66.0 & \underline{76.7} & \underline{63.0} & 59.7 & 54.5 & \underline{70.8}  &  \underline{97.5} & \underline{68.5} & \underline{79.5} & 72.0 & 43.8 &  63.3     \\
        {\fontsize{8}{10}\selectfont Gemini-1.5-Pro} & \underline{67.4} & \underline{74.3} & \underline{66.7}& 76.5 &  62.8 & 59.0 & \underline{65.0}  & 61.0 &  92.5 & 53.5 & 49.5 & 59.4 & 50.0 & 61.2   \\
        {\fontsize{8}{10}\selectfont Gemini-2.0-flash} & 56.3 & 65.7 & 51.3 & 58.3 & 50.7 & 50.0 & 62.0 & 48.2 & 82.0 & 39.5 & 43.0 & 39.0 & 49.6 & 36.2    \\
        {\fontsize{8}{10}\selectfont Gemini-1.5-flash} & 46.1 &  49.5 &  45.2 & 60.2 & 48.3 & 32.2 & 41.5  &46.8 & 79.0   & 33.0 & 50.0 & 41.2 & 55.5 & 22.0      \\

        \addlinespace[2pt]
        \midrule
        \addlinespace[2pt]
        
        \multicolumn{15}{c}{ \textit{Open-Source MLLMs} }   \\ \midrule
         {\fontsize{8}{10}\selectfont Llama-3.2-90B-Vision} & 37.6 & 43.7 & 37.3 & 49.2 & 35.3 & 36.0 & 24.0 & 50.6 &  \underline{94.5} & 32.5 & 53.0 & 39.7  & 59.6 & 24.3 \\
         {\fontsize{8}{10}\selectfont Llama-3.2-11B-Vision} & 19.7 & 29.7 & 13.0 & 25.7 & 28.7 &9.3 & 12.0 & 33.2  &  72.0 & 23.8 & 36.5 & 16.2  & 39.7 & 11.2   \\
       {\fontsize{8}{10}\selectfont InternVL2\_5-78B} & 41.0 & 42.3 & 35.3 & 43.3 & 35.7 & 40.3 & 49.0 & 55.2 & 82.0  & 43.0 & 59.0 & 63.9  & 45.1 & 38.2   \\
          {\fontsize{8}{10}\selectfont InternVL2\_5-38B} & 31.3 & 37.3 & 33.0 & 38.3 & 25.3 &17.3 & 36.5 & 44.0 & 61.5 & 32.5 & 49.0 & 39.5 & 46.3 & 35.0  \\
        {\fontsize{8}{10}\selectfont InternVL2\_5-8B} & 2.0 & 4.0 &6.0 &2.0 & 0.0 & 0.0 & 0.0 & 19.4 & 40.2 & 11.5 & 11.0  & 16.0 & 30.7 & 7.3 \\
       {\fontsize{8}{10}\selectfont Qwen2-VL-72B-Ins} & 38.7 & 45.3 & 33.3 &44.7 & 35.7 &33.0 & 40.0  & 42.2 & 72.0  & 33.0 & 39.0 & 37.0 & 52.0 &  20.3   \\
       {\fontsize{8}{10}\selectfont Qwen2-VL-7B-Ins} & 5.2 & 8.3 & 5.3 & 7.0 & 1.7 &3.3 & 5.5 & 26.1 &  53.3 & 9.0 & 24.0 & 25.8 & 41.2& 3.2  \\
        \midrule
        \addlinespace[2pt]
        \multicolumn{15}{c}{ \textit{Proprietary MLLMs with BrainMem} } \\
        \midrule
    
        {\fontsize{8}{10}\selectfont GPT-4o}
        &  79.2 & 84.3 & 80.7 & 83.2 & 66.7 & \textbf{80.3} & 79.7
        & 73.9 & 93.3 & 60.0 & 70.2 & \textbf{79.3} & \textbf{64.1} & \textbf{76.2}\\
    
        {\fontsize{8}{10}\selectfont Claude-3.5-Sonnet}
        &  75.7 & 78.3 & 72.0 & \textbf{84.7} & 68.2 & 66.7 & \textbf{84.3}
        &  \textbf{76.2} & \textbf{96.7} & \textbf{74.3}& \textbf{85.5} & 77.7 & 53.2 & 70.0 \\
    
        {\fontsize{8}{10}\selectfont Gemini-1.5-pro}
        &  \textbf{80.1} & \textbf{92.3} & \textbf{84.7} & 81.7& \textbf{76.2} & 75.1 & 70.3
        &  68.3 & 95.5 & 60.7 & 61.5 & 69.3 & 55.3 & 67.2\\

        \bottomrule
    \end{tabular}
    }

\end{table*}

\subsection*{A.2 Average Steps Analysis}

Tables~\ref{tb:high_env_steps_suppl} and \ref{tab:low_env_steps_suppl} report average planner steps and environment steps across the four benchmarks.
We focus on the comparison between the original planners and their BrainMem-augmented versions.

\paragraph{Memory consistently reduces environment steps.}
Across all four benchmarks, adding BrainMem leads to fewer environment interactions, indicating more stable execution and lower action redundancy.
For example, in EB-Habitat, Claude-3.5-Sonnet drops from 12.1 to \textbf{10.4} env steps, and GPT-4o decreases from 13.1 to 12.1.
Similar reductions appear in EB-Manipulation, where GPT-4o improves from 12.9 to \textbf{12.1} env steps, showing that memory reduces unnecessary exploration and repeated corrections.

\paragraph{Memory slightly increases planner steps (as expected).}
Planner steps occasionally increase (e.g., GPT-4o in EB-ALFRED from 4.4 to 4.8), since BrainMem supplies richer contextual information, leading to more deliberate planning.
However, the increase is small (typically +0.2–0.4), and the resulting plans require **fewer physical interactions**, yielding a favorable trade-off.

\paragraph{Episodic recall and spatial consistency reduce wandering.}
Benchmarks requiring spatial reasoning—EB-Navigation and EB-Habitat—show clear benefits:
Claude-3.5-Sonnet and GPT-4o reduce env steps to \textbf{15.2} and \textbf{12.1}, respectively.
This reflects BrainMem’s ability to prevent repeated room scanning and maintain object-awareness even after objects temporarily leave the field of view.

\paragraph{Manipulation benefits from temporal stability.}
In EB-Manipulation, all models show fewer env steps when equipped with memory (e.g., Gemini-1.5-Pro: 13.4 → 12.9).
This indicates that memory helps avoid incorrect object poses, unstable grasps, or premature releases—issues that cause unnecessary corrective actions in memory-less planners.

\begin{table}[t]
    \centering
    \renewcommand{\arraystretch}{1.15}
    \caption{\textbf{Average planner and environment steps} for EB-ALFRED and EB-Habitat.}
    
\centering
\scriptsize
\setlength{\tabcolsep}{3pt}
\renewcommand{\arraystretch}{1.05}
\label{tb:high_env_steps_suppl}
\begin{tabular}{l|cc|cc}
\toprule
\textbf{Model} 
& \multicolumn{2}{c|}{\textbf{EB-ALFRED}} 
& \multicolumn{2}{c}{\textbf{EB-Habitat}} \\
& Plan & Env & Plan & Env \\
\midrule
GPT-4o                & 4.4 & 16.3 & 5.5 & 13.1 \\
GPT-4o-mini           & 7.7 & 20.6 & 7.4 & 18.8 \\
Claude-3.5-Sonnet     & 4.0 & 12.1 & 4.2 & 10.9 \\
Gemini-1.5-Pro        & 3.9 & 15.7 & 5.4 & 12.6 \\
Gemini-2.0-flash      & 4.4 & 16.3 & 6.8 & 14.8 \\

GPT-4o(with BrainMem)                & 4.8 & 14.3 & 6.2 & 12.1 \\
Claude-3.5-Sonnet(with BrainMem)     & 4.6 & \textbf{10.9} & 4.9 & \textbf{10.4} \\
Gemini-1.5-Pro(with BrainMem)        & 4.5 & 13.8 & 5.8 & 11.7 \\
\bottomrule
\end{tabular}

    \label{tab:high_env_steps_suppl}
     \vspace{-0.5cm}
\end{table}

\begin{table}[t]
    \centering
    \renewcommand{\arraystretch}{1.15}
    \caption{\textbf{Average planner and environment steps} for EB-Navigation and EB-Manipulation.}
    
\centering
\scriptsize
\setlength{\tabcolsep}{3pt}
\renewcommand{\arraystretch}{1.05}

\begin{tabular}{l|cc|cc}
\toprule
\textbf{Model} 
& \multicolumn{2}{c|}{\textbf{EB-Navigation}} 
& \multicolumn{2}{c}{\textbf{EB-Manipulation}} \\
& Plan & Env & Plan & Env \\
\midrule
GPT-4o                & 6.2 & 15.5 & 2.6 & 12.9 \\
GPT-4o-mini           & 7.6 & 17.5 & 3.4 & 14.7 \\
Claude-3.5-Sonnet     & 6.2 & 15.6 & 2.7 & 13.3 \\
Gemini-1.5-Pro        & 8.8 & 16.5 & 2.7 & 13.4 \\
Gemini-2.0-flash      & 9.2 & 16.0 & 2.8 & 14.0 \\
GPT-4o(with BrainMem)                & 6.4 & \textbf{15.2} & 3.2 & \textbf{12.1} \\
Claude-3.5-Sonnet(with BrainMem)     & 6.2 & \textbf{15.2} & 3.2 & 12.7 \\
Gemini-1.5-Pro(with BrainMem)        & 8.6 & 15.4 & 3.1 & 12.9 \\
\bottomrule
\end{tabular}

    \label{tab:low_env_steps_suppl}
    \vspace{-0.5cm}
\end{table}

\subsection*{A.3 Extra Experimental Settings}

For completeness and reproducibility, we document the experimental setup used for all proprietary planners and benchmarks. All models are evaluated under a unified inference configuration to ensure consistency. Specifically, every model is run with a temperature of 0 and a maximum completion length of 2048 tokens. All visual observations are standardized to 500×500 resolution before being passed to the planner.

We adopt the default multi-step planning controller across all tasks, with environment interaction budgets aligned to task complexity:
30 steps for high-level EB-ALFRED and EB-Habitat tasks,
20 steps for EB-Navigation,
and 15 steps for EB-Manipulation.
These limits prevent runaway loops while keeping evaluation comparable across models.

All benchmarks follow the official EmbodiedBench splits, including the predefined training subsets and held-out test episodes. During evaluation, we use task success rate as the primary metric, since it directly reflects the agent’s ability to complete the goal specification under each simulator’s action space and environmental constraints.

\subsection*{A.4 Robustness}
\begin{table}[t]
\centering
\small
\setlength{\tabcolsep}{6pt}

\caption{Robustness to 5 random episode orders on Base Set.}

\label{tab:rebuttal_shuffle}
{
\resizebox{0.5\textwidth}{!}{
\begin{tabular}{lccccc}
\toprule
Shuffle Order & 1 & 2 & 3 & 4 & 5 \\
\hline
SR (\%) & 80 & 82 & 82 & 80 & 84 \\
\hline
Mean $\pm$ Std & \multicolumn{5}{c}{$81.6 \pm 1.7$} \\
\bottomrule
\end{tabular}}
}

\end{table}

To assess robustness to ordering effects, we tested BrainMem across five randomized episode orders on EB-ALFRED (Table~\ref{tab:rebuttal_shuffle}). Results show consistently stable performance with low variance.

\vspace{0.5em}

\newtcblisting{codebox}[2][]{
  listing only,
  colback=gray!3,
  colframe=blue!70,
  coltitle=white,
  fonttitle=\sffamily\bfseries\small,
  title=#2,
  boxrule=0.8pt,
  arc=1pt,
  left=4pt,
  right=4pt,
  top=4pt,
  bottom=4pt,
  breakable,
  enhanced jigsaw,
  before skip=8pt,
  after skip=8pt,
  listing options={
    basicstyle=\scriptsize\ttfamily,
    breaklines=true,
    breakatwhitespace=false,
    columns=flexible,
    keepspaces=true,
    showstringspaces=false
  },
  #1
}

\newtcolorbox{examplebox}[2][]{
  colback=gray!3,
  colframe=blue!70,
  coltitle=white,
  fonttitle=\sffamily\bfseries\small,
  title=#2,
  boxrule=0.8pt,
  arc=1pt,
  left=4pt,
  right=4pt,
  top=4pt,
  bottom=4pt,
  breakable,
  enhanced jigsaw,
  before skip=8pt,
  after skip=8pt,
  #1
}

\section*{B. Extended Method Details}

\subsection*{B.1 Memory Contents in EB-ALFRED}

We provide concrete examples of how each memory component in \textbf{BrainMem} is instantiated and updated on EB-ALFRED tasks:

\begin{itemize}[leftmargin=1.2em,noitemsep,topsep=3pt]
    \item \textbf{Working Memory}: short-term execution context over recent steps,
    \item \textbf{Episodic Memory}: action-transition graphs for completed attempts,
    \item \textbf{Spatial Memory}: room-object and object-object relations,
    \item \textbf{Semantic Memory}: distilled high-level experiences and guidelines.
\end{itemize}

Each subsection below shows the raw stored structure and, when applicable, the prompt-facing format sent to the planner.

\subsubsection*{B.1.1 Working Memory Example}

Working memory maintains a sliding window (size $W$) over the most recent interaction steps in the current episode. Each step record contains (conceptually):
\[
[\text{Action},\ \text{Feedback},\ \text{Agent State},\ \text{Location},\ \text{Timestamp}]
\]

An example internal entry at step $t=3$:

\begin{codebox}{Working Memory Internal Structure}
{
  "step_index": 3,
  "action": "MoveAhead",
  "feedback": "success",
  "reasoning": "Approaching the sink to clean the fork",
  "observation_summary": "Sink visible at 2m distance, fork in hand",
  "agent_state": {"holding": "fork"},
  "location": "kitchen",
  "timestamp": 3
}
\end{codebox}

When serialized to the planner prompt, the working memory window is rendered as a concise textual history:

\begin{examplebox}{Working Memory Prompt Format}
\small
Working Memory (last 3 steps)

[Step 1] Action: PickUp(fork), Feedback: success\\
\hspace*{1em}Agent holding: fork, Location: kitchen

[Step 2] Action: MoveAhead, Feedback: success\\
\hspace*{1em}Agent holding: fork, Location: near sink

[Step 3] Action: MoveAhead, Feedback: success\\
\hspace*{1em}Agent holding: fork, Location: in front of sink

Current short-term success rate: 100\%
\end{examplebox}

This prompt segment keeps the planner grounded in recent actions, outcomes, and agent state, reducing repeated failures and short-sighted loops.

\subsubsection*{B.1.2 Episodic Memory (Trajectory KG)}

Episodic memory stores each completed attempt (success or failure) as a directed \emph{Trajectory KG}. Nodes correspond to abstracted states; edges correspond to actions with outcomes.

\paragraph{State node.}
A state node $s_i$ is stored as:
\[
s_i = (\text{state\_id},\ \text{room\_id},\ \text{summary})
\]
where \texttt{summary} is a short text description (e.g., visible objects, agent pose, key relations).

\paragraph{Transition edge.}
A transition $e_{i \to j}$ is stored as:
\[
e_{i \to j} = (\text{prev\_state\_id},\ \text{next\_state\_id},\ \text{action},\ \text{success},\dots)
\]

\noindent\textbf{Example graph snippet (clean fork task):}

\begin{examplebox}{Trajectory Knowledge Graph Structure}
\small
State s\_001: ``Fork on countertop, sink visible''

\hspace*{1em}--[PickUp(fork), success=True]--$>$

State s\_002: ``Holding fork, 2m from sink''

\hspace*{1em}--[MoveAhead, success=True]--$>$

State s\_003: ``Holding fork, in front of sink''

\hspace*{1em}--[PutObject(sink), success=True]--$>$

State s\_004: ``Fork placed in sink, ready to clean''
\end{examplebox}

Across episodes, Trajectory KGs are used to retrieve high-value or informative trajectories conditioned on task type, room, and key objects.

\subsubsection*{B.1.3 Spatial Memory (Room-Object Relations)}

Spatial memory aggregates stable layout patterns across episodes in the same scene/room. A spatial entry records which objects appear in a room and their relations:

\begin{codebox}{Spatial Memory Structure}
{
  "scene_id": "FloorPlan1",
  "room_id": "kitchen",
  "objects": ["sink", "countertop", "fork", "cabinet"],
  "relations": [
    {"source": "sink", "relation": "near", "target": "countertop"},
    {"source": "fork", "relation": "on", "target": "countertop"}
  ],
  "support_count": 12
}
\end{codebox}

Conceptually, this corresponds to the relations:
\[
(\text{room} \xrightarrow{\text{contains}} \text{object}_i),\quad
(\text{object}_i \xrightarrow{\text{near/on/in}} \text{object}_j).
\]

These spatial patterns are queried during planning to answer questions such as ``where is the sink likely located in this kitchen?'' or ``where are knives usually found?''.

\subsubsection*{B.1.4 Semantic Memory (Experiences and Guidelines)}

Semantic memory stores distilled experiences and guidelines derived from completed episodes.

\paragraph{Experience entry.}
Each episode is summarized into a compact experience record:

\begin{codebox}{Experience Record Structure}
{
  "experience_id": "exp_042",
  "task_instruction": "Put a clean fork on the dining table",
  "room_id": "kitchen",
  "success": true,
  "action_summary": "PickUp(fork) -> MoveTo(sink) -> PutObject(sink)",
  "key_failure_reasons": [],
  "key_success_pattern": "Always place the fork into the sink 
before turning on the faucet to clean it."
}
\end{codebox}

\paragraph{Guideline entry.}
From multiple compatible experiences, a reusable guideline is formed:

\begin{codebox}{Guideline Structure}
{
  "guideline_id": "guide_clean_sink_01",
  "task_type": "clean",
  "object_category": "dish",
  "description": "Before turning on the faucet, ensure the 
target object is placed in the sink.",
  "success_count": 12,
  "total_usage": 14,
  "utility_score": 0.91
}
\end{codebox}

During planning, we retrieve top-$k$ guidelines matching the current task type and object category and inject them into the planner prompt as high-level advice.

\subsection*{B.2 Episodic KG Construction Details}

\subsubsection*{B.2.1 Node and Edge Formation}

At the beginning of an episode, BrainMem creates an initial state node $s_0$ summarizing the initial scene and task instruction. For each environment step $t$:

\begin{enumerate}[leftmargin=1.5em,noitemsep,topsep=3pt]
    \item Observe the environment and summarize key objects and agent pose.
    \item Create or reuse a state node $s_t$ with this summary.
    \item Execute the chosen action $a_t$ and record its outcome.
    \item Create a transition edge $e_{t-1 \to t}$ linking $s_{t-1}$ to $s_t$.
\end{enumerate}

\noindent Example evolution for a cleaning task:

\begin{examplebox}{Episodic KG Evolution}
\small
Step 0: s\_000 = ``Fork on countertop, sink visible''

Step 1: s\_000 --[PickUp(fork)]--$>$ s\_001

Step 2: s\_001 --[MoveAhead]--$>$ s\_002

Step 3: s\_002 --[PutObject(sink)]--$>$ s\_003

Step 4: s\_003 --[TurnOnFaucet]--$>$ s\_004
\end{examplebox}

\subsubsection*{B.2.2 Merging Across Episodes}

When multiple episodes share similar states or transitions, we merge them to form a denser, more informative KG:

\begin{itemize}[leftmargin=1.2em,noitemsep,topsep=3pt]
    \item \textbf{State deduplication}: states with the same room and summary reuse the same node ID.
    \item \textbf{Transition aggregation}: transitions with identical (prev\_state, action, next\_state) accumulate success counts and attach multiple reasoning examples.
    \item \textbf{Failure paths}: failed transitions are tagged and later recalled when the same task is retried, helping the agent avoid known bad actions.
\end{itemize}

These rules allow episodic memory to grow with experience while remaining queryable and compact.

\subsection*{B.3 Semantic Memory Utility Scoring}

To prevent semantic memory from drifting or being dominated by outdated heuristics, we maintain a lightweight \emph{utility score} for each guideline.

\subsubsection*{B.3.1 Statistics and Scores}

For each guideline $g$, we track:
\[
N_{\text{success}}(g),\quad N_{\text{total}}(g)
\]
and define its \emph{confidence} as:
\[
\text{confidence}(g) = \frac{N_{\text{success}}(g)}{N_{\text{total}}(g)}.
\]

We also define a bounded usage term:
\[
\text{usage}(g) = \min\left(\frac{N_{\text{success}}(g)}{10},\ 1.0\right),
\]
and combine them into a composite utility:
\[
\text{utility}(g) = 0.7 \cdot \text{confidence}(g) + 0.3 \cdot \text{usage}(g).
\]

After each episode where $g$ is applied, we update the counts and recompute its utility.

\subsubsection*{B.3.2 Pruning Low-Utility Guidelines}

Semantic memory keeps at most $M$ active guidelines (e.g., $M=20$). If the number of guidelines exceeds a soft upper limit (e.g., $1.5M$), we sort by utility and prune the lowest-scoring entries. High-confidence, frequently successful guidelines (e.g., $\text{confidence} \geq 0.8$ and $N_{\text{success}} \geq 5$) are protected from pruning.

This simple mechanism keeps $\mathcal{M}_\text{sem}$ both compact and adaptive: new, useful patterns are quickly promoted, while stale or unreliable rules are gradually removed.

\vspace{-0.5em}
\newtcolorbox{promptbox}[2][]{
  colback=gray!3,
  colframe=blue!70,
  coltitle=white,
  fonttitle=\sffamily\bfseries\small,
  title=#2,
  boxrule=0.8pt,
  arc=1pt,
  left=4pt,
  right=4pt,
  top=4pt,
  bottom=4pt,
  breakable,
  enhanced jigsaw,
  before skip=8pt,
  after skip=8pt,
  #1
}

\section*{C. Prompt Templates and Examples}

\subsection*{C.1 Working Memory Prompt Example}

Working memory is formatted as a sliding window of recent action steps, including action outcomes, reasoning, and observation summaries. Example prompt:

\begin{promptbox}{Working Memory Prompt}
\small
\textbf{Enhanced Memory Context:}

\textbf{Working Memory (Recent Actions):}\\
Step 1: Action: pick up the Fork (Success, reward=0.10)\\
\hspace*{1em}Observation Summary: Fork visible on countertop\\
\hspace*{1em}Reasoning: Need to pick up fork first\\
Step 2: Action: find a Sink (Success, reward=0.05)\\
\hspace*{1em}Observation Summary: Moving towards sink\\
\hspace*{1em}Reasoning: Navigate to sink for cleaning\\
Step 3: Action: put down the Fork (Success, reward=0.15)\\
\hspace*{1em}Observation Summary: Sink visible, fork in hand\\
\hspace*{1em}Reasoning: Placing fork in sink for cleaning\\
Step 4: Action: turn on the Faucet (Failure, reward=0.00)\\
\hspace*{1em}Observation Summary: Cannot turn on faucet without object in sink\\
\hspace*{1em}Reasoning: Attempted to turn on faucet but fork not properly placed

\textbf{Key Guidelines:}
\begin{itemize}[leftmargin=1.2em,noitemsep,topsep=3pt]
\item Check current robot state above (what you are holding)
\item Do not repeat failed actions immediately
\item Trust environment feedback (``Object is in [Location]'' indicates direct navigation)
\item For cleaning: Pick up, Put in sink, Turn faucet on/off, Pick up from sink
\item For complex placement: Pick up small object, Put in container, Pick up container, Place
\item If stuck seeing same objects, move to different rooms
\end{itemize}

\textbf{CONSISTENCY WARNING}: Cannot turn on the Faucet - you are not holding any object. For water-based cleaning: (1) Pick up object, (2) Put in sink, (3) Turn on/off faucet. Please carefully reconsider your action choice based on recent history.

\textbf{Recently Visited Locations}: kitchen, countertop, sink

\textbf{CURRENT ROBOT STATE:}
\begin{itemize}[leftmargin=1.2em,noitemsep,topsep=3pt]
\item Currently holding: \textbf{NOTHING}
\item CRITICAL: Always check this before planning your next action
\item If holding something and need to pick up another object, MUST put down first
\end{itemize}

\textbf{Performance}: Success Rate 75.0\%, Actions 4

\textbf{Strategy}: Use working memory to avoid failures, apply room patterns, reference experiences.
\end{promptbox}

Each step includes: (1) step index and action taken, (2) success/failure status with reward, (3) observation summary describing the visual state, and (4) the LLM's reasoning for that action. The prompt also includes consistency warnings for invalid action sequences, recently visited locations, current robot state (holding status), and performance metrics.

\subsection*{C.2 Episodic Retrieval Prompt Example}

Episodic KG retrieval converts graph structures into textual summaries for planner prompts. The retrieval includes multiple components:

\textbf{Room Successful Patterns:} Task-specific action sequences that succeeded in the same room:

\begin{promptbox}{Room Successful Patterns}
\small
\textbf{Room Successful Patterns:}
\begin{enumerate}[leftmargin=1.5em,noitemsep,topsep=3pt]
\item clean: pick up the Fork; find a Sink; put down the Fork; turn on the Fauce
\item clean: pick up the Spoon; find a Sink; put down the Spoon; turn on the Faucet
\end{enumerate}
\end{promptbox}

\textbf{Real-time Action Hints:} Object-specific successful actions retrieved from the episode KG:

\begin{promptbox}{Real-time Action Hints}
\small
\textbf{Successful actions for fork:} pick up the Fork; find a Sink; put down the Fork
\end{promptbox}

\textbf{Spatial Reasoning Guidance:} General spatial reasoning principles:

\begin{promptbox}{Spatial Reasoning Guidance}
\small
\textbf{Spatial Reasoning:}
\begin{itemize}[leftmargin=1.2em,noitemsep,topsep=3pt]
\item Observe the current view to understand object positions
\item Note spatial relationships: left/right, front/back, near/far
\item Consider which objects to approach first based on their locations
\item Plan efficient movement routes between objects
\end{itemize}
\end{promptbox}

These retrieved components are combined with working memory and semantic guidelines to provide comprehensive context for decision-making.

\subsection*{C.3 Semantic Guideline Prompt Example}

Semantic guidelines are high-confidence strategies extracted from successful experiences. They are provided in full (no retrieval needed) and prioritized in decision-making:

\begin{promptbox}{Semantic Guidelines}
\small
\textbf{Valuable Guidelines (Proven Strategies):}\\
These guidelines have been validated through successful task completions. PRIORITIZE them in your decision making:

\begin{enumerate}[leftmargin=1.5em,topsep=4pt]
\item Always pick up objects before navigating to cleaning locations. Check sink availability before attempting PutObject.\\
(Validated 23 times, 92\% confidence) [clean, place]

\item Navigate to sink before PutObject. Ensure object is in hand before moving to sink area.\\
(Validated 18 times, 90\% confidence) [clean]

\item For container-object manipulation: first locate container, then pick up object, then navigate to container, finally PutObject near container.\\
(Validated 15 times, 88\% confidence) [place, container]

\item For cleaning tasks: Pick up object, Put in sink, Turn on faucet, Rinse, Turn off faucet, Pick up, Place at target.\\
(Validated 12 times, 85\% confidence) [clean, rinse, wet]

\item For complex placement (``set X with Y in it''): Pick up small object (Y), Put into container (X), Pick up container (X), Place at target.\\
(Validated 10 times, 83\% confidence) [place, container]
\end{enumerate}

\textbf{CRITICAL}: These guidelines come from proven successful experiences. Follow them whenever applicable.
\end{promptbox}

Each guideline includes: (1) the strategy text, (2) success count and confidence percentage, and (3) applicable task types. Guidelines are sorted by utility score (confidence $\times$ 0.7 + usage $\times$ 0.3) and the top $M=20$ are included in every prompt. Examples include cleaning/heating routines, container-object manipulation strategies, and multi-step navigation hints.

\subsection*{C.4 Experience Agent Prompt Example}

The Experience Agent analyzes complete episode logs to extract semantic experiences. This happens after episode completion in \texttt{experience\_agent.py}:

\begin{promptbox}{Experience Agent Analysis Prompt}
\footnotesize
You are an expert AI assistant analyzing robot task execution logs. Your job is to extract valuable semantic experiences from episode execution data.

\textbf{Analysis Context:}
\begin{itemize}[leftmargin=1.2em,noitemsep,topsep=3pt]
\item Robot operates in household environments (kitchen, living room, bedroom, bathroom)
\item Tasks involve object manipulation, navigation, and complex multi-step sequences
\item Robot can only hold one object at a time
\item ``Put down'' near containers automatically places objects INSIDE containers
\end{itemize}

\textbf{Your Task:}\\
Analyze the provided episode execution log and extract:

\begin{enumerate}[leftmargin=1.5em,noitemsep,topsep=3pt]
\item \textbf{Success Pattern} (if episode succeeded):
\begin{itemize}[leftmargin=1.2em,noitemsep]
\item Key successful action sequences
\item Critical decision points that led to success
\item Effective navigation patterns
\item Successful object interaction strategies
\end{itemize}

\item \textbf{Failure Analysis} (if episode failed):
\begin{itemize}[leftmargin=1.2em,noitemsep]
\item Root causes of failure
\item Specific actions or decisions that led to failure
\item Alternative approaches that could have worked
\item Common mistakes to avoid
\end{itemize}

\item \textbf{Learning Insights} (always):
\begin{itemize}[leftmargin=1.2em,noitemsep]
\item General principles learned from this episode
\item Room-specific strategies discovered
\item Object interaction patterns identified
\item Navigation efficiency improvements
\end{itemize}
\end{enumerate}

\textbf{Output Format:}\\
Provide a JSON response with:

\begin{lstlisting}[basicstyle=\scriptsize\ttfamily,breaklines=true]
{
  "episode_success": true/false,
  "primary_task": "brief task description",
  "success_patterns": ["list of successful strategies"],
  "failure_causes": ["list of failure root causes"],
  "learning_insights": ["list of general insights"],
  "action_sequences": {
    "successful": ["key successful action patterns"],
    "failed": ["action patterns that failed"]
  },
  "recommendations": ["specific recommendations"]
}
\end{lstlisting}

Be concise but specific. Focus on actionable insights that can improve future performance.
\end{promptbox}

\textbf{Streaming KG Manager Prompts:} The \texttt{enhanced\_streaming\_kg\_manager.py} uses simplified prompts for quick experience extraction at episode end.

For successful tasks:

\begin{promptbox}{Success Experience Extraction}
\small
Based on the following successfully completed task information, extract key successful experiences and patterns:

Task Instruction: [task instruction]\\
Room: [room\_id]\\
Task Type: [task\_type]\\
Action Sequence: [action sequence]\\
Number of Steps: [step count]

Please concisely summarize the key experiences from this success, focusing on:
\begin{enumerate}[leftmargin=1.5em,noitemsep,topsep=3pt]
\item Effective action strategies
\item Important spatial layout utilization
\item Key points in object interaction
\end{enumerate}

Experience Summary (1-2 sentences):
\end{promptbox}

For failed tasks:

\begin{promptbox}{Failure Reflection Extraction}
\small
Based on the following failed task information, conduct a reflective analysis:

Task Instruction: [task instruction]\\
Room: [room\_id]\\
Task Type: [task\_type]\\
Action Sequence: [action sequence]\\
Failure Reason: Task not completed

Please analyze the failure reasons and extract learning points:
\begin{enumerate}[leftmargin=1.5em,noitemsep,topsep=3pt]
\item Possible strategy issues
\item Areas that need improvement
\item Behaviors to avoid next time
\end{enumerate}

Reflection Summary (1-2 sentences):
\end{promptbox}

\subsection*{C.5 Full Planner Prompt Demonstration}

A complete end-to-end prompt integrating all memory components during a test episode:

\begin{promptbox}{Full Planner Prompt}
\footnotesize
\textbf{SYSTEM PROMPT}\\
(Base ALFRED system prompt with action descriptions and guidelines)

\textbf{Streaming Memory System:}\\
You have multi-layer memory: working memory (last 5 actions), episodic memory (knowledge graphs indexed by room number and task type), and semantic memory (experience and guideline).

\textbf{Core Rules:}
\begin{enumerate}[leftmargin=1.5em,noitemsep,topsep=3pt]
\item \textbf{Working Memory}: Always record last 5 actions and current holding status before acting
\item \textbf{Single Object}: Robot can only hold one object at a time
\item \textbf{Container Logic}: ``Put down'' near a container automatically places object INSIDE container
\item \textbf{Episodic Graph}: Store object-location-action transitions per room number, predict state changes
\item \textbf{Experience Storage}: Record task, strategy, outcome, and cause for success/failure reflection
\item \textbf{Guideline Application}: Apply rules for tasks that are often done incorrectly or inefficiently
\end{enumerate}

\textbf{Cleaning Tasks} (rinse/clean/wet):\\
Pick up object, Put in sink, Turn on faucet, Rinse, Turn off faucet, Pick up, Place at target

\textbf{TASK INSTRUCTION}\\
Put a clean fork on the dining table.

\textbf{MEMORY CONTEXT}

\textbf{Enhanced Memory Context:}

\textbf{Valuable Guidelines (Proven Strategies):}
\begin{enumerate}[leftmargin=1.5em,noitemsep,topsep=3pt]
\item Always pick up objects before navigating to cleaning locations. [Validated 23 times, 92\% confidence] [clean, place]
\item Navigate to sink before PutObject. [Validated 18 times, 90\% confidence] [clean]
\end{enumerate}

\textbf{Room Successful Patterns:}
\begin{enumerate}[leftmargin=1.5em,noitemsep,topsep=3pt]
\item clean: pick up the Fork; find a Sink; put down the Fork
\end{enumerate}

\textbf{Relevant Experiences:}
\begin{enumerate}[leftmargin=1.5em,noitemsep,topsep=3pt]
\item Always pick up objects before navigating to cleaning locations. Check sink availability before attempting PutObject.
\item Navigate to sink area first, then PutObject when close enough.
\end{enumerate}

\textbf{Successful actions for fork:} pick up the Fork; put down the Fork

\textbf{Spatial Reasoning:}
\begin{itemize}[leftmargin=1.2em,noitemsep,topsep=3pt]
\item Observe the current view to understand object positions
\item Note spatial relationships: left/right, front/back, near/far
\item Consider which objects to approach first based on their locations
\item Plan efficient movement routes between objects
\end{itemize}

\textbf{Working Memory (Recent Actions):}\\
(Step 1) Action: pick up the Fork (Success, reward=0.10)\\
\hspace*{1em}Observation Summary: Fork visible on countertop\\
\hspace*{1em}Reasoning: Following guideline to pick up object first\\
(Step 2) Action: find a Sink (Success, reward=0.05)\\
\hspace*{1em}Observation Summary: Moving towards sink\\
\hspace*{1em}Reasoning: Navigating to sink for cleaning

\textbf{Key Guidelines:}
\begin{itemize}[leftmargin=1.2em,noitemsep,topsep=3pt]
\item Check current robot state above (what you are holding)
\item Do not repeat failed actions immediately
\item For cleaning: Pick up, Put in sink, Turn faucet on/off, Pick up from sink
\end{itemize}

\textbf{CURRENT ROBOT STATE:}
\begin{itemize}[leftmargin=1.2em,noitemsep,topsep=3pt]
\item Currently holding: \textbf{FORK}
\item CRITICAL: Always check this before planning your next action
\end{itemize}

\textbf{Performance}: Success Rate 100.0\%, Actions 2

\textbf{Strategy}: Use working memory to avoid failures, apply room patterns, reference experiences.

\textbf{CURRENT OBSERVATION}\\
(Visual observation: RGB image, object detection...)

\textbf{ACTION SPACE}\\
find, pick up, put down, turn on, turn off, open, close, slice, ...

\textbf{YOUR RESPONSE}\\
Provide: (1) reasoning, (2) next action
\end{promptbox}

The prompt structure follows this order: (1) Valuable Guidelines (highest priority, always included), (2) Room Successful Patterns (episodic KG retrieval), (3) Relevant Experiences (semantic memory), (4) Real-time Action Hints (episodic KG), (5) Spatial Reasoning Guidance, and (6) Working Memory (recent action history with consistency warnings and current robot state). This hierarchical organization ensures the planner prioritizes validated strategies while incorporating contextual and recent information

\vspace{-0.5em}

\section*{D. Additional Visualizations}

\subsection*{D.1 Visualization of Three Additional Tasks}
\begin{figure*}[t]
    \centering
    \includegraphics[width=\linewidth]{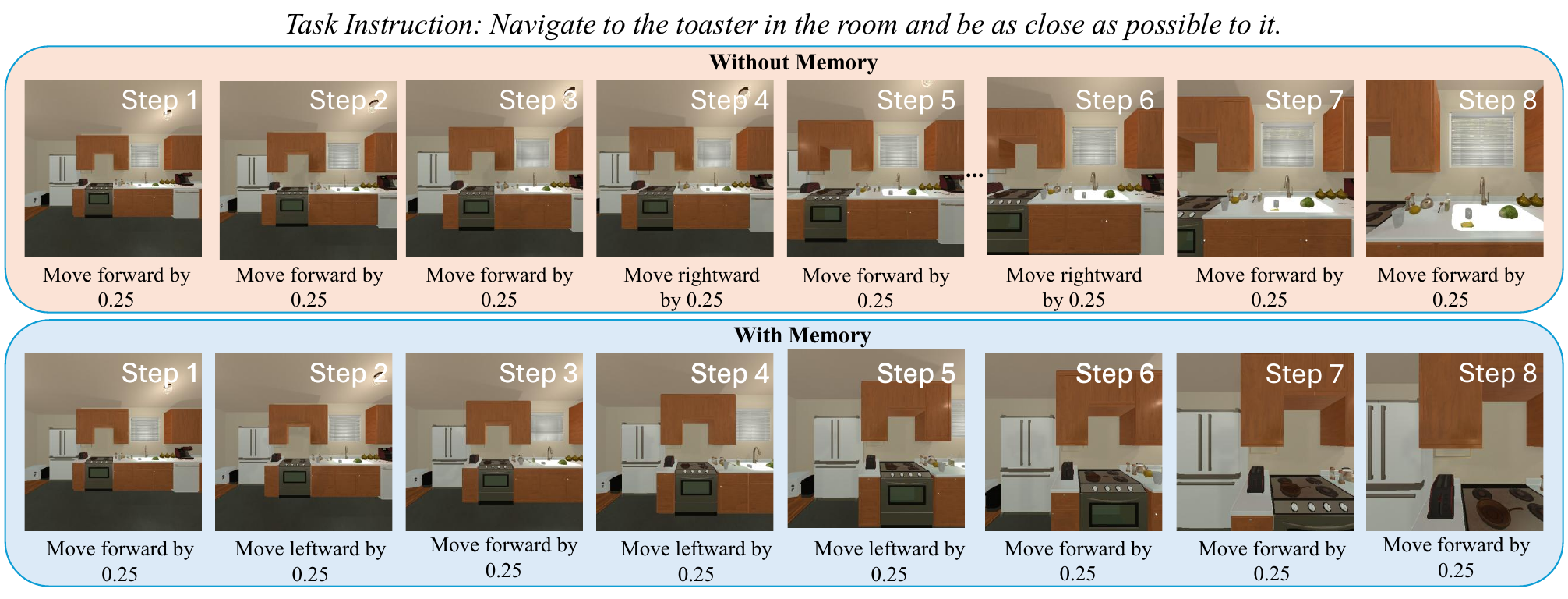}
    \caption{\textbf{Navigation task visualization (without vs. with memory).}
    Task: “Navigate to the toaster in the room and be as close as possible to it.”  
    Without memory, the agent sees the toaster in the first four steps but repeatedly moves rightward, increasing the distance instead of approaching it. This reflects a lack of persistent spatial grounding. With memory, the agent maintains awareness of the toaster's location and navigates directly toward it, successfully reaching the target.}
    \label{fig:nav_toaster}
\end{figure*}

\begin{figure*}[t]
    \centering
    \includegraphics[width=\linewidth]{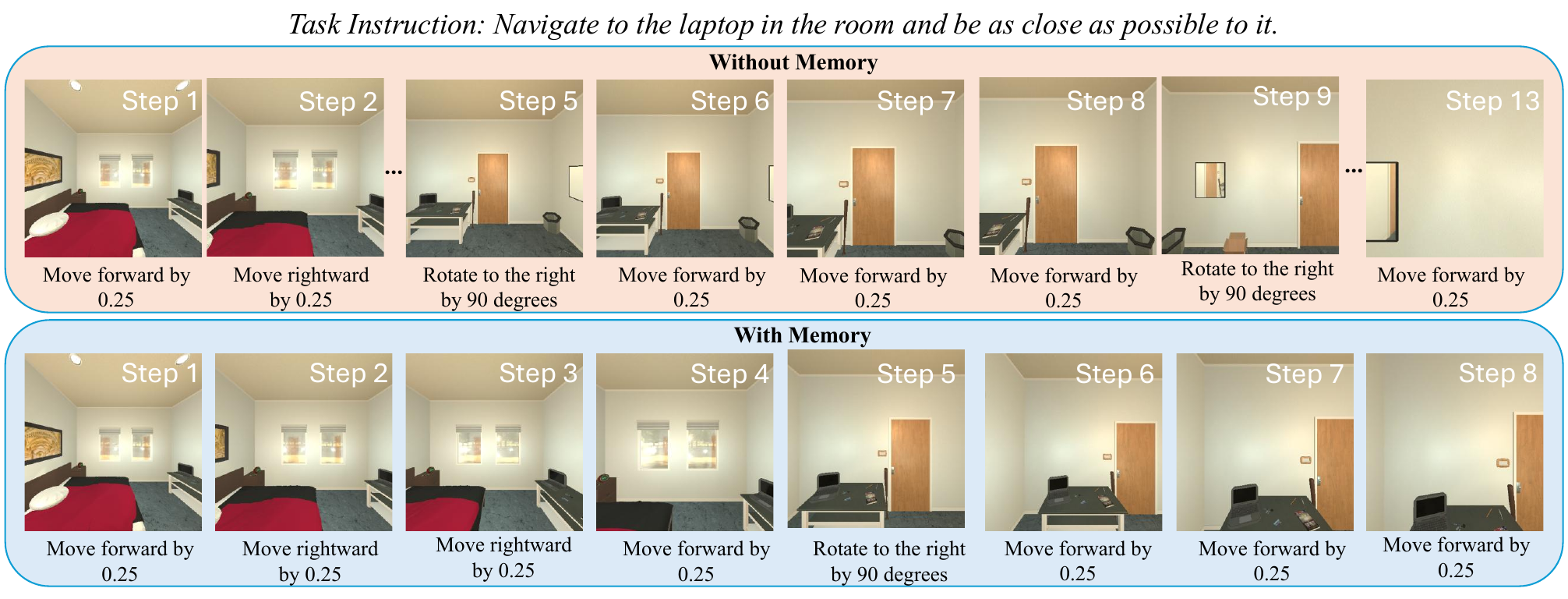}
    \caption{\textbf{Navigation task visualization (without vs. with memory).}
    Task: “Navigate to the laptop in the room and be as close as possible to it.”  
    Without memory, although the laptop is visible early, the agent moves forward in a way that narrows the field of view and loses track of the laptop, showing weak spatial depth reasoning. With memory, the agent retains knowledge of the laptop's position and adjusts its trajectory accordingly, reaching the goal reliably.}
    \label{fig:nav_laptop}
\end{figure*}

We include additional paired visualizations (without vs. with memory) for: EB-Habitat, EB-Navigation, EB-Manipulation.
Across the six additional qualitative visualizations, we observe consistent failure modes in memory-less planners. In the toaster navigation task (Fig.~\ref{fig:nav_toaster}), the agent keeps moving toward the far right of the room despite the toaster being visible for several consecutive steps, increasing its distance from the target instead of approaching it. In the laptop navigation task (Fig.~\ref{fig:nav_laptop}), although the laptop is initially visible, the agent loses track of it after stepping forward; the narrowing field of view hides the object, and without recalling previous observations, the agent drifts without reorienting toward the goal.

\begin{figure*}[t]
    \centering
    \includegraphics[width=\linewidth]{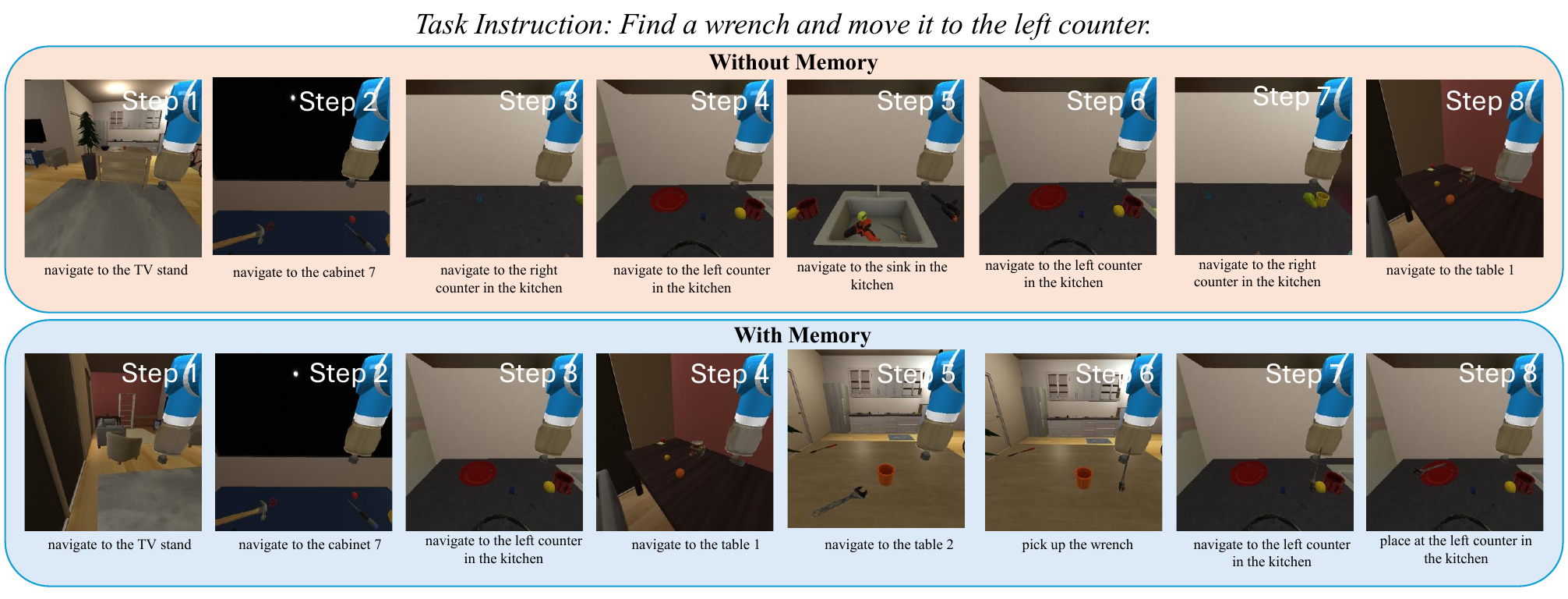}
    \caption{\textbf{Habitat task visualization (without vs. with memory).}
    Task: “Find a wrench and move it to the left counter.”  
    Without memory, the agent repeatedly re-searches previously visited regions, failing to maintain a record of explored areas. This leads to inefficient looping behavior and missing the wrench despite previously observing it. With memory, the agent recalls explored zones and retrieved spatial cues, enabling efficient search and successful delivery to the counter.}
    \label{fig:hab_wrench}
\end{figure*}

\begin{figure*}[t]
    \centering
    \includegraphics[width=\linewidth]{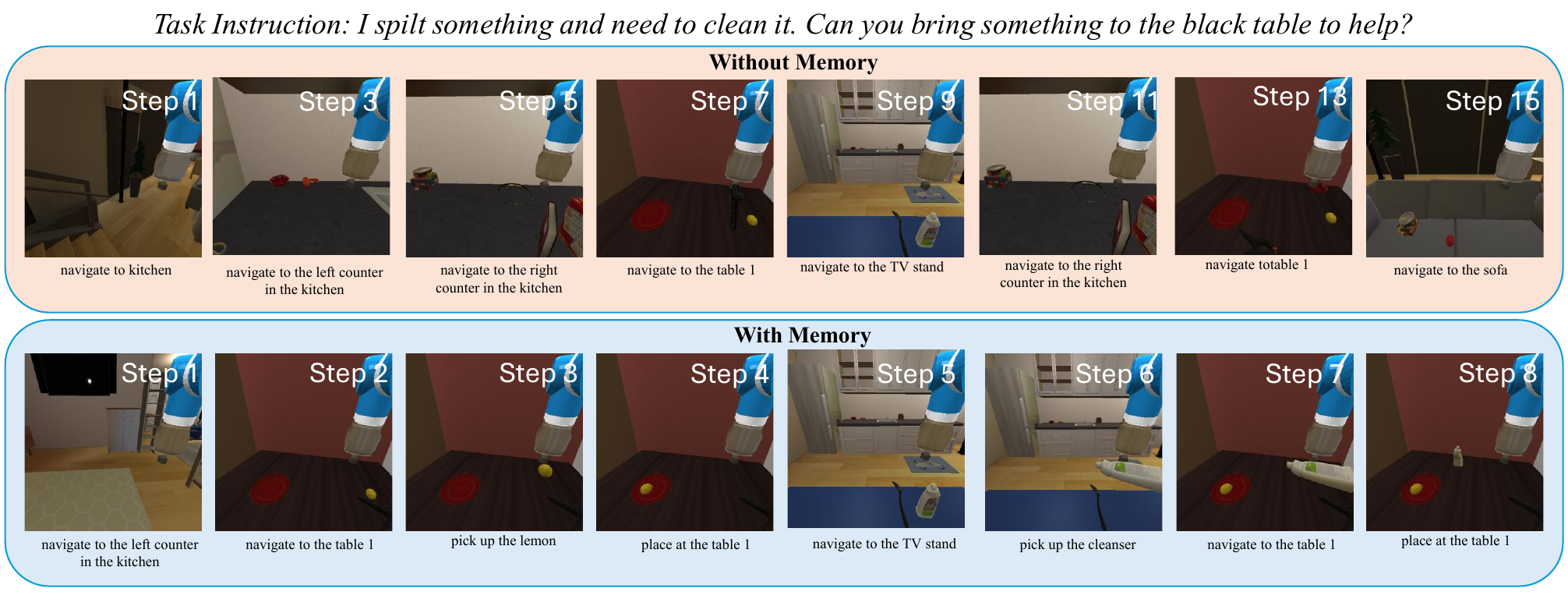}
    \caption{\textbf{Habitat task visualization (without vs. with memory).}
    Task: “I spilled something and need to clean it. Bring something to the black table to help.”  
    Without memory, the agent briefly sees the cleaning tool (cleanser) but fails to track its relevance and instead wanders into unrelated rooms, losing sight of the correct object and ultimately failing. With memory, the agent retains the observation of the cleanser and consistently pursues the correct sequence, successfully retrieving and placing the cleaning item at the target location.}
    \label{fig:hab_clean}
\end{figure*}

In the Habitat manipulation tasks, the absence of memory leads to additional inefficiencies. In the wrench search task (Fig.~\ref{fig:hab_wrench}), the memory-less agent repeatedly revisits previously searched regions, wasting substantial steps and failing to maintain global search coverage. In the cleaning task (Fig.~\ref{fig:hab_clean}), the agent directly sees the correct cleanser early in the episode but ignores it and continues exploring unrelated areas, ultimately causing task failure.

\begin{figure*}[t]
    \centering
    \includegraphics[width=\linewidth]{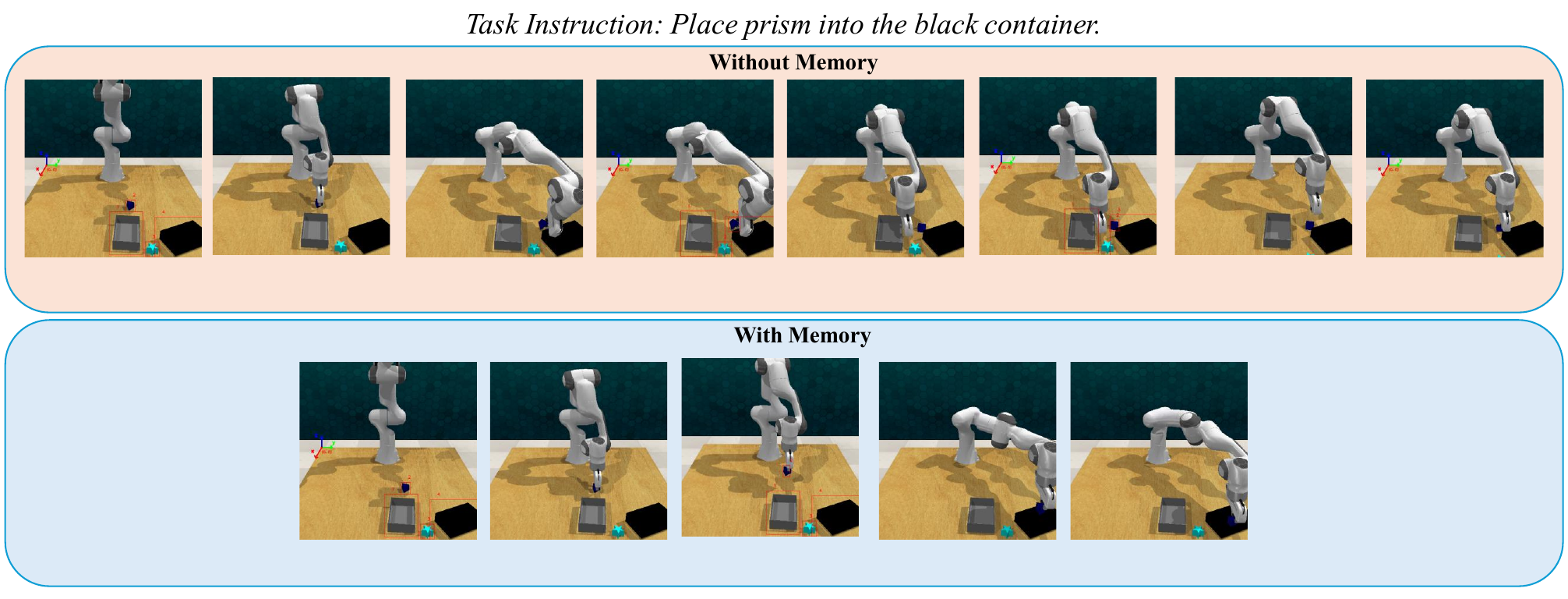}
    \caption{\textbf{Manipulation task visualization (without vs. with memory).}
    Task: “Place the prism into the black container.” 
    \textit{Without memory}, the agent misjudges the physical constraints and repeatedly collides the prism against the rim of the black container, causing the object to bounce away and leading to failure. This reflects a lack of physical interaction awareness and no recall of prior unsuccessful attempts. 
    \textit{With memory}, the agent recalls stable placement trajectories and aligns the prism carefully before lowering it smoothly into the container, completing the task successfully.}
    \label{fig:man_prism}
\end{figure*}

\begin{figure*}[t]
    \centering
    \includegraphics[width=\linewidth]{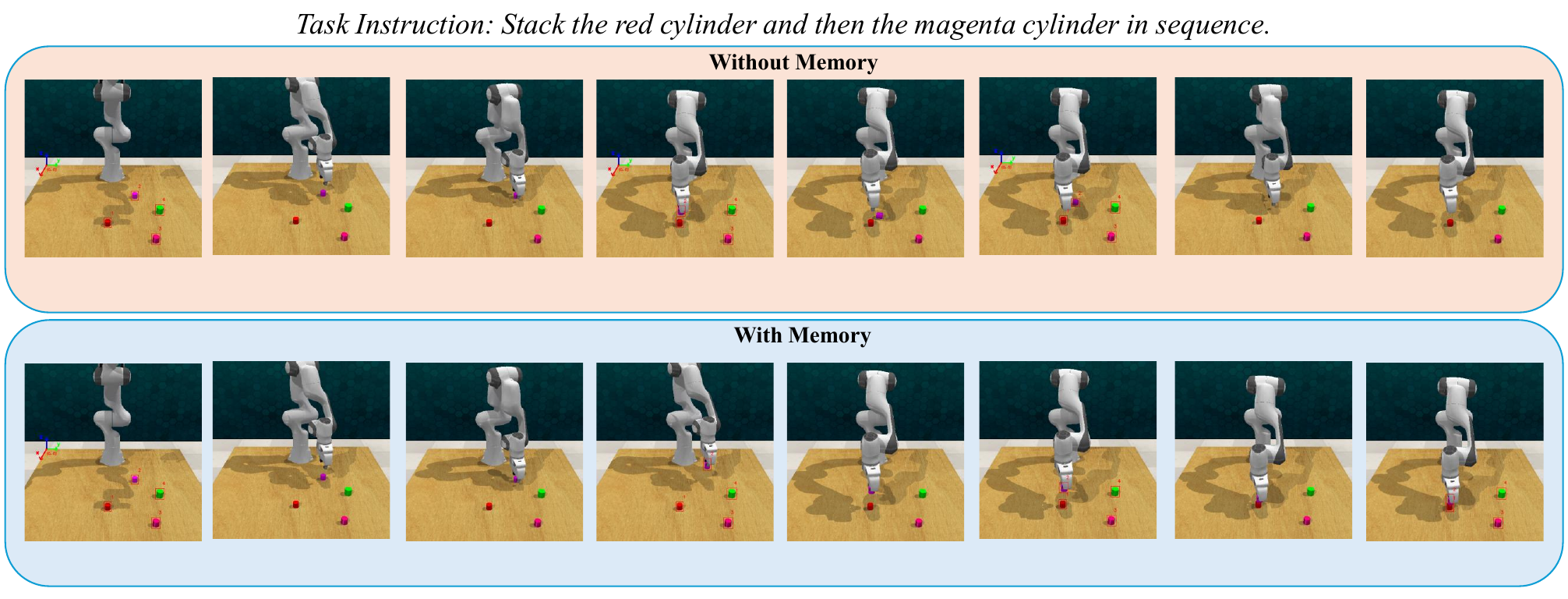}
    \caption{\textbf{Manipulation task visualization (without vs. with memory).}
    Task: “Stack the red cylinder and then the magenta cylinder in sequence.” 
    \textit{Without memory}, the agent releases objects prematurely without accounting for vertical alignment—dropping cylinders from a distance instead of gently lowering them—resulting in unstable stacking and repeated failures.
    \textit{With memory}, the agent retrieves previous successful stacking behaviors, aligns the red cylinder first, then lowers the magenta cylinder smoothly, achieving a stable two-level stack.}
    \label{fig:man_stack}
\end{figure*}

For physical object manipulation, the lack of temporal and spatial recall results in poor interaction stability. In the ``Place the prism into the black container'' task (Fig.~\ref{fig:man_prism}), the agent repeatedly collides with the container rim due to an inability to infer the contact geometry from previous failed attempts. In the stacking task (Fig.~\ref{fig:man_stack}), the agent releases cylinder too early rather than slowly lowering it into position, causing unstable placement and immediate collapse.

In contrast, the memory-augmented BrainMem agent succeeds in all six tasks. 
Working memory preserves short-term continuity, ensuring that visible targets, spatial cues, and action outcomes remain accessible across steps. Episodic memory provides structured recall of successful past trajectories—such as safe object approach angles or efficient room traversal paths—allowing the agent to avoid repeating known failures. Semantic memory supplies distilled, task-general rules (e.g., ``lower objects before releasing'', ``prioritize visible target objects'', ``avoid re-exploring visited areas''), enabling robust and transferable planning strategies. Together, these complementary memory modules yield coherent, consistent, and failure-resistant behavior that memory-less planners consistently fail to produce.

\end{document}